\documentclass[conference]{IEEEtran}
\IEEEoverridecommandlockouts

\usepackage{graphicx}
\usepackage[misc,geometry]{ifsym} 

\usepackage[utf8]{inputenc} 
\usepackage[T1]{fontenc}    
\usepackage{hyperref}       

\usepackage{url}            
\usepackage{booktabs}       
\usepackage{amsfonts}       
\usepackage{nicefrac}       
\usepackage{microtype}      

\usepackage{times}
\usepackage{soul}
\usepackage{url}
\usepackage{caption}

\usepackage{amsthm,amsmath}
\usepackage{booktabs}
\usepackage{algorithm}
\usepackage[noend]{algpseudocode}
\usepackage{tikz}
\usetikzlibrary{positioning}
\usepackage{pgfplots}
\usepackage{microtype}
\usepackage[noend]{algpseudocode}
\usepackage{subcaption}
\graphicspath{ {Results/} }

\usepackage{enumitem, hyperref}
\makeatletter
\def\namedlabel#1#2{\begingroup
	#2%
	\def\@currentlabel{#2}%
	\phantomsection\label{#1}\endgroup
}
\makeatother

\usepackage{multicol}
\usepackage{multirow}

\usepackage{nccmath}
\usepackage{mathtools}
\usepackage{bbm}
\pgfplotsset{compat=1.16}

\def\beq{\begin{equation}}\def\eeq{\end{equation}}\usepackage[T1]{fontenc}

\usepackage{cleveref}

\usepackage[switch]{lineno}

\usepackage{todonotes}

{}

\begin{document}

\title{
Impact Of Missing Data Imputation On The Fairness And Accuracy Of Graph Node Classifiers}


\author{\IEEEauthorblockN{Haris Mansoor$^{*1}$, Sarwan Ali$^{*2}$, Shafiq Alam$^{3}$, Muhammad Asad Khan$^{4}$, Umair ul Hassan$^{5}$, Imdadullah Khan$^{1+}$}
\IEEEauthorblockA{
Lahore University of Management Science, Lahore, Pakistan$^{1}$\\
Georgia State University, Atlanta, USA$^{2}$ \\
Massey Business School, Massey University, Auckland, New Zealand$^{3}$ \\
Hazara University, Mansehra, Pakistan$^{4}$\\
National University of Ireland Galway, Galway, Ireland $^{5}$ \\
16060061@lums.edu.pk, sali85@student.gsu.edu, 
S.Alam1@massey.ac.nz \\
asadkhan@hu.edu.pk, umair.ulhassan@nuigalway.ie, imdad.khan@lums.edu.pk} 
\\
{$^{*}$ Equal Contribution} 
{$^{+}$ Corresponding Author}
\thanks{2022 IEEE International Conference on Big Data (Big Data) | 978-1-6654-8045-1/22/\$31.00 \copyright 2022 European Union}
}

%
%

\maketitle
\thispagestyle{plain}
\pagestyle{plain}
\begin{abstract}
Analysis of the fairness of machine learning (ML) algorithms recently attracted many researchers' interest. Most ML methods show bias toward protected groups, which limits the applicability of ML models in many applications like crime rate prediction etc. Since the data may have missing values which, if not appropriately handled, are known to further harmfully affect fairness. Many imputation methods are proposed to deal with missing data. However, the effect of missing data imputation on fairness is not studied well. In this paper, we analyze the effect on fairness in the context of graph data (node attributes) imputation using different embedding and neural network methods. Extensive experiments on six datasets demonstrate severe fairness issues in missing data imputation under graph node classification. We also find that the choice of the imputation method affects both fairness and accuracy. Our results provide valuable insights into graph data fairness and how to handle missingness in graphs efficiently. This work also provides directions regarding theoretical studies on fairness in graph data.



\end{abstract}

\begin{IEEEkeywords}
Graphs, Fairness, Experimental Study, Bias, GNNs, Equal Opportunity, Demographic Parity
\end{IEEEkeywords}


\section{Introduction}\label{section:introduction}
Many real-world datasets have some form of graph structure associated with them: social networks~\cite{dai2021say}, paper citations~\cite{giles1998citeseer}, protein-protein interactions~\cite{hamilton2017inductive}, etc. Numerous machine learning algorithms are proposed to deal particularly with the graph data (e.g., Graph Neural Networks (GNNs)~\cite{hamaguchi2017knowledge}, Node2Vec~\cite{grover2016node2vec}, DeepWalk~\cite{perozzi2014deepwalk}, and Graph Summarization~\cite{ali2021ssag}). These algorithms have shown remarkable improvement in results on graph data~\cite{louizos2015variational,tang2020transferring,tang2020investigating}. Most of them are designed to integrate both node features, and graph edges information (topological structure) which enhance node representation and improve results~\cite{yao2019graph,ying2018graph,zhao2020semi}. Although these algorithms can capture graph structure information successfully, the social bias in data can cause fairness issues~\cite{Rahman2019FairwalkTF}. This limits the applicability of most of the algorithms in practical graph data applications.

Many studies regarding fairness in machine learning models have reported that a lot of datasets contain discrimination and social bias towards a sensitive attribute like region, age, skin color, gender, etc.~\cite{beutel2017data,creager2019flexibly,dwork2012fairness}. Machine learning models trained on such data can inherit bias. Recently many studies on the fairness of GNNs and Node2Vec have also reported similar results~\cite{buyl2020debayes,khajehnejad2022crosswalk,Rahman2019FairwalkTF}. As stated earlier, most of these algorithms are designed to incorporate edge information, which provides extra data patterns and better representation. But the bias in data can also propagates through the edges too. This can aggravate the fairness issue in graphs~\cite{grover2016node2vec,perozzi2014deepwalk}.

Another important behavior in graph data is the presence of homophily~\cite{ali2021predicting} (connections among nodes having the same attribute values). 
In social network graphs, nodes with similar sensitive features tend to connect with each other as compared to nodes with different sensitive features.  For example, older people tend to have friends in a similar age group. In such cases, aggregating neighbor node features produces representations of similar age groups quite different from that of other age groups, leading to severe bias in graph-specific algorithms. Recent research demonstrated that such results tend to correlate with sensitive attribute~\cite{beutel2017data,dwork2012fairness}. Missing data is common in every machine learning problem, and if not dealt with properly, has an adverse effect on fairness~\cite{bakker2020fair}.

\begin{figure}[h!]
    \centering
    \begin{subfigure}{.25\textwidth}
  \centering
  \includegraphics[scale=0.3] {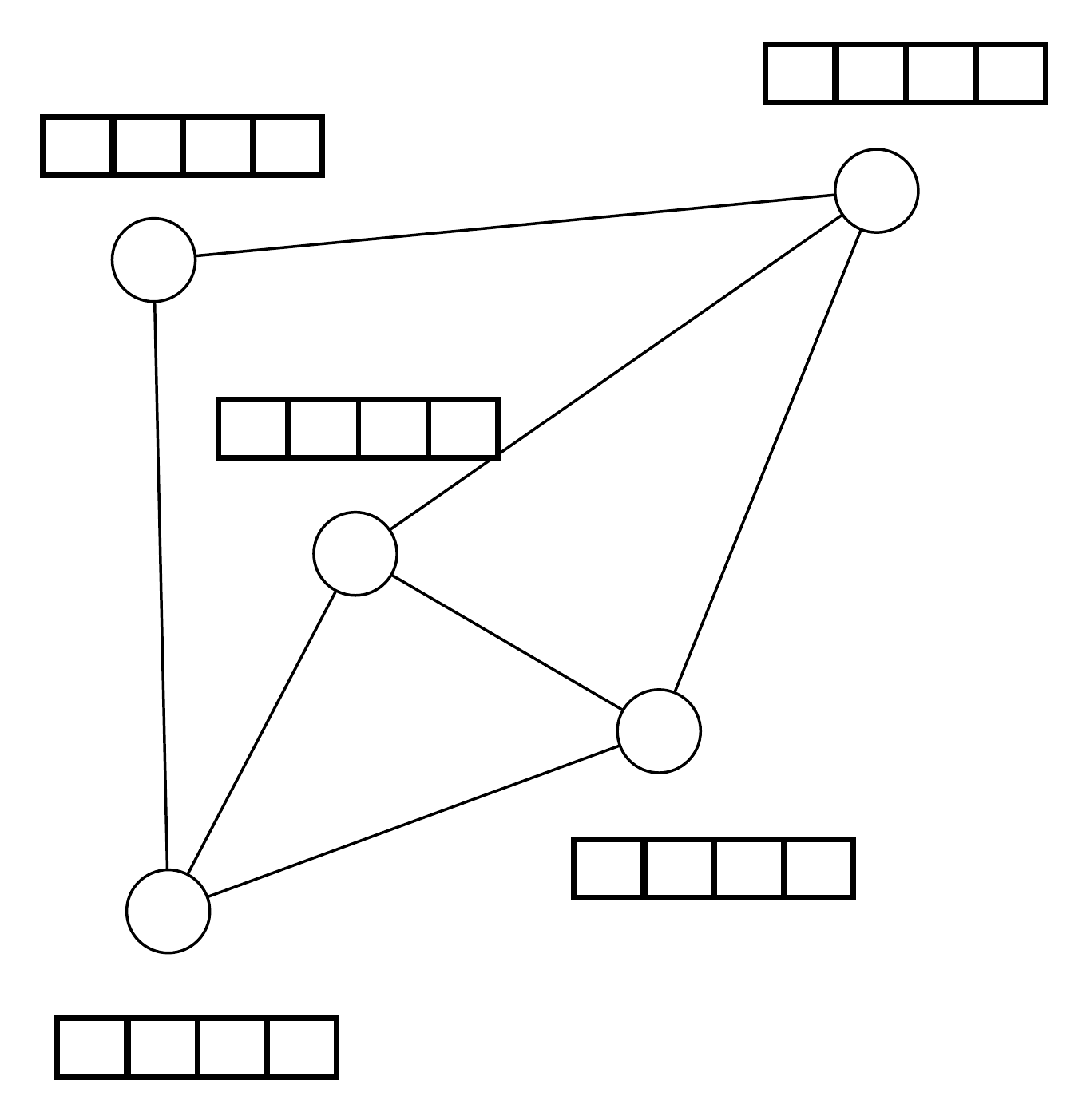}
  \caption{}
  \label{fig:graph_without_missing}
\end{subfigure}%
\begin{subfigure}{.25\textwidth}
  \centering
  \includegraphics[scale=0.3] {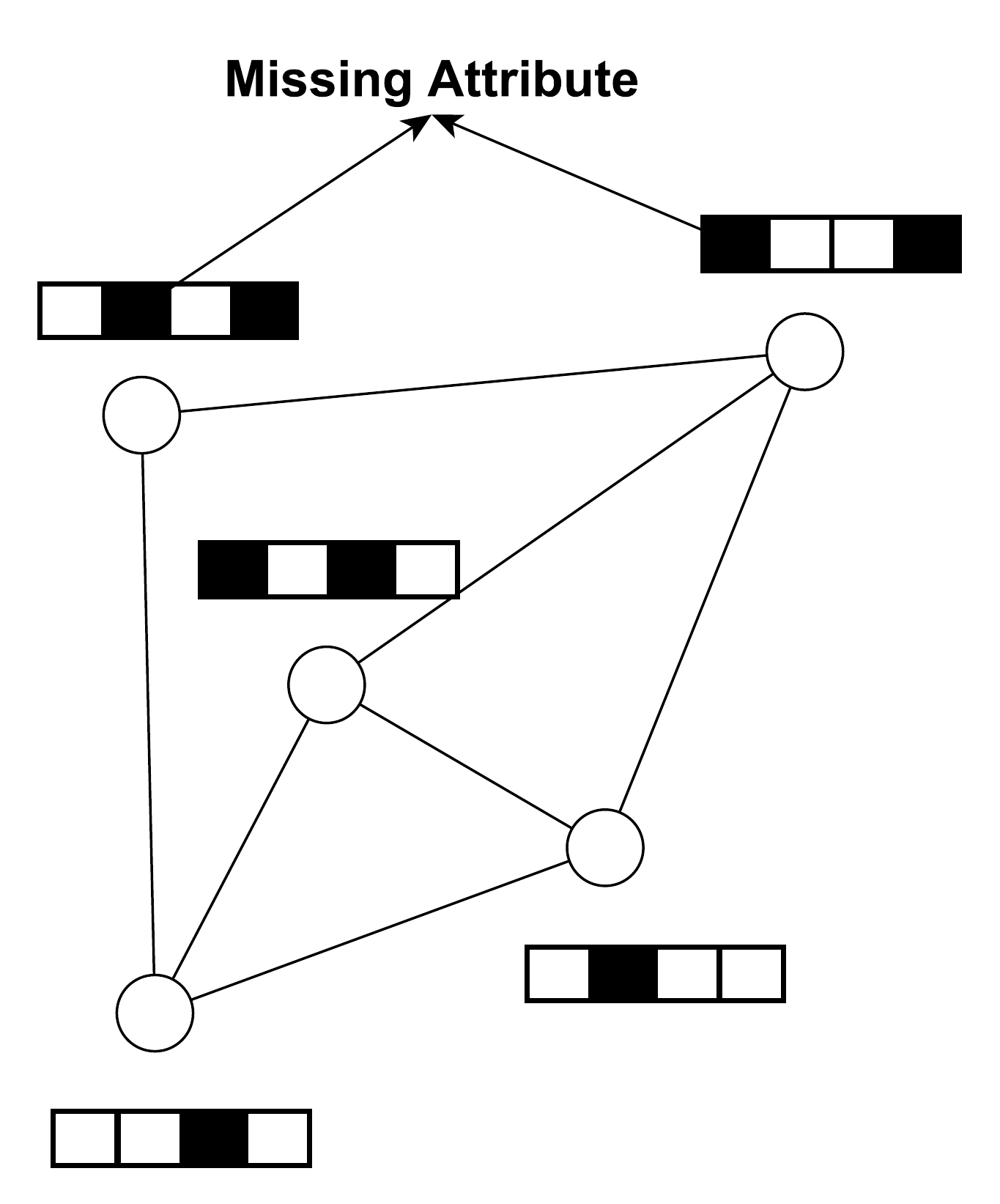}
  \caption{}
  \label{fig:graph_with_missing}
\end{subfigure}%
\caption{A graph containing five nodes, each node has four features. (a) all features are fully observed, (b) black squares represent missing features.}  
    \label{fig_graph_missing}
\end{figure}

The most common approach to dealing with missing data is imputation (see Figure~\ref{fig_graph_missing}). Different statistical and machine learning imputation methods have been proposed to deal with missing data~\cite{lovedeep2018multiple,li2019misgan,yang2019categorical}. Many researchers discussed the main cause of unfairness is the bias in the dataset~\cite{beutel2017data,creager2019flexibly,edwards2015censoring,hardt2016equality}. Missing values can further cause sensitive attribute imbalance. This implies that missing data can decrease fairness. There is limited research work regarding the impact of missing data imputation on fairness. Recently some studies discussed that missing data contribute to bias in machine learning algorithms~\cite{bakker2020fair,martinez2019fairness,rajkomar2018ensuring}. Zhang et al.~\cite{Fairness_in_missing_data} investigate the impact of missing data imputation on fairness and conclude that missing data imputation produces a bias in the data. However, they do not include graph data in the studies. Another work that addresses the problem of accessing fairness under missing data is performed in~\cite{zhang2021assessing}. They also reported that missing data produces challenges in a classifier's fairness.

Motivated by the recent advancements in graph classification algorithms (GNNs~\cite{hamaguchi2017knowledge}, Node2Vec~\cite{grover2016node2vec}, DeepWalk~\cite{perozzi2014deepwalk}) and fairness studies, in this work we investigate the impact of missing data imputation on the fairness of graph node classification problem. Specifically, we perform an extensive experimental study using six datasets and different imputation methods to empirically evaluate the effect of missing data imputation on fair oblivious and fair node classifiers.  

To our knowledge, no work has been performed regarding the fairness of data imputation on graph structure data. In this context, our work differs from~\cite{Fairness_in_missing_data} as they have only studied fairness on tabular data imputation. Our work is also different from~\cite{zhang2021assessing} as they only study fairness under missing data without considering imputation. Both of these works do not consider graph structure data.

\subsection{Our Contribution:}

To the best of our knowledge, we are the first to study the fairness of missing data imputation on graph data. Our empirical studies show that:
\begin{itemize}
    \item Fairness in graph data is affected by missing data imputation.
    \item Data imputation methods have an impact on fairness and accuracy. 
    \item Missing data mechanisms have an adverse effect on fairness. 
    \item Most fairness issues are associated with sample imbalance.
    \item Missing data also affect the accuracy of the classifiers. 
    \item Fair node classifiers improve bias but at the cost of accuracy.
    \item Fairness under missing data imputation is subject to the specific algorithm used.
    \item A trade-off between accuracy and fairness widely exists in the results.
\end{itemize}

The rest of the paper is organized as follows. Related work is reviewed in Section~\ref{sec:related_work}. In Section~\ref{sec:preliminaries}, the preliminaries are explained. The node classifiers used in the paper are explained in Section~\ref{sec:node class}. We provide the detail regarding experimental setup in Section~\ref{sec:experimental_setup}. The results are reported in Section~\ref{sec:results}. Finally, we conclude the paper in Section~\ref{sec:conclusions}.

\section{Related Work} \label{sec:related_work}

In this section, we review the research related to fairness in graph data. 
Specifically, we summarize the literature on node classification in graph data, fairness in node classification, and fairness in data imputation.    

\subsection{Graph Node Classifiers}
Machine learning problems involving graph data are mainly solved by representing data as node embeddings (generated from nodes and edges) and then training the classifier algorithm on embeddings. 
DeepWalk~\cite{perozzi2014deepwalk} is a commonly used algorithm to generate vector representations of nodes in a graph. It generates low-dimensional representations for nodes in a graph through random walks. It has shown promising results in many graph learning problems, including node classification.
Proposed in~\cite{grover2016node2vec}, Node2Vec is another well-known algorithm that uses a biased walk (breadth first, depth first) instead of a random walk. It exhibits particular promise for a variety of machine learning applications~\cite{grover2016node2vec,khosla2019comparative}.

Recently, \emph{graph neural networks} (GNNs) have been proposed to combine node embedding and classification tasks using single neural networks. Substantial improvements in results have been reported using GNNs when compared to other node classification techniques. 
The GNNs can be divided into two types: spectral-based GNNs~\cite{bruna2013spectral,defferrard2016convolutional,henaff2015deep}  and spatial-based GNNS~\cite{chiang2019cluster,hamilton2017inductive,velickovic2017graph}.
Graph Convolutional Networks (GCNs) is an example of the spectral-based method as they implement the convolution operation on graph data~\cite{kipf2016semi}. Spatial-based GNNs are motivated by aggregating neighbor nodes' information to find a node representation~\cite{hamilton2017inductive}. Graph attention networks (GATs) integrate the self-attention mechanism in spatial aggregation such that it assigns higher weights to more important nodes~\cite{velickovic2017graph}. Both GCNs and GATs have shown great results on node classification tasks.

\subsection{Fairness in Graph Node Classification}

Although, fairness in machine learning is a well-studied problem~\cite{beutel2017data,creager2019flexibly,edwards2015censoring,hardt2016equality}. Little attention has been paid to the issue of fairness in graph node classification~\cite{garg2020fairness}. Extant literature explores this issue in graph node embeddings or node classifiers.

The first group of research works focuses on the design of an embedding mechanism that would be fair for the sensitive attribute~\cite{buyl2020debayes,khajehnejad2022crosswalk,Rahman2019FairwalkTF}. Most of these works pertain to improving Node2Vec~\cite{grover2016node2vec} or DeepWalk~\cite{perozzi2014deepwalk} algorithms by creating fair embeddings which target specific applications of graph data.

The second group of research works aims to design an end-to-end solution for a fair node classification problem. These solutions can take care of embeddings and classification simultaneously. For instance, Dai et al.~\cite{dai2021say} proposed a fair node classifier called FairGNN. It can work on data with limited sensitive attribute information. The basic idea is to design a loss function that incorporates fairness in the problem. 

\subsection{Fairness in Data Imputation }
Existing research on fairness in data imputation is mainly limited to tabular data. Zhang et al.~\cite{Fairness_in_missing_data} studied various imputation techniques and reported that data imputation could potentially introduce more bias in machine learning models. 
They conclude that data imputation further harms the imbalance of sensitive attributes, causing unfairness (e.g., if the males have more missingness as compared to the females). 
Another work that addresses the problem of accessing fairness under missing data is presented  in~\cite{zhang2021assessing}. In this work, authors trained machine learning models on fully observed data (with completely observed features) and then tested them on missing data. They reported that missing data  produced challenges in the fairness of classifiers and showed that a model which is trained using only a subset of the whole dataset (complete case domain) might be fair for the chosen subset but not for the whole dataset (complete data domain). 
Compared to the above research, we focus on the case of graph data and study the impact of data imputation on the fairness of node classifiers.

\section{Preliminaries}\label{sec:preliminaries}
In this section, we first define the notation used throughout the paper. Then, we explain different notions of fairness used in the literature. We also introduce the methods used to introduce missingness in data and summarize the existing techniques for data imputation.
\subsection{Notation}

Assume an attributed graph that has attributes associated with nodes. For example, social networks contain attributes of individuals. 
Let $G = (V, E, X, Y)$ denote an attributed graph, where $V=\{v_1,v_2,...,v_N \}$ is the set of $N$ nodes, and $E \subseteq V \times V $ is the set of edges between them. There is a set of features $X=\{x_1,x_2,...,x_N \}$ and labels $Y=\{y_1,y_2,...,y_N \}$ associated with $V$. Such that each node $v_i$ has a feature vector $x_i\in \mathbb{R^K} $, and a binary label  $y_i\in \{+,- \} $. There is a binary sensitive attribute $A \in \{a,b\}$ in the dataset, where $a$ is the majority group and $b$ is the minority group. The problem is to predict node labels (represented by $P$), such that the algorithm supposes not to discriminate results based on $A$.

\subsection{Measures of Fairness}
The fairness problem in machine learning algorithms is generally assessed based on a predefined sensitive attribute $A$. Our focus is on binary-sensitive attributes and binary node classification problems. The majority and minority groups in $A$ are represented by $a$ and $b$, respectively. Since machine learning algorithms make decisions based on the available data, the bias in data propagates in results, causing fairness issues. 

We use \emph{demographic parity} and \emph{equal opportunity} as fairness measures. Both of these are widely used by the research community to assess the fairness of machine learning models in graph data~\cite{agarwal2021towards,buyl2020debayes,dai2021say,dong2022edits,fan2021fair,khajehnejad2022crosswalk,zhang2021multi}. For a binary sensitive attribute $A\in\{a,b\}$, and binary class label $y\in\{+,-\}$, such that $TP$, $TN$, $FP$, $FN$ represents the true positive, true negative, false positive, false negative with respect to $a$ or $b$ (as stated), the fairness measures are defined below.

	

\begin{itemize}
    \item Demographic (statistical) parity is defined as the probability of being assigned to a positive class is independent of the sensitive attribute, i.e., $P(+|a) = P(+|b) \ \ \forall a,b \in A$. This requires the acceptance rates of the candidates from both groups to be equal, thus introducing fairness with respect to sensitive attributes. We use the parity difference between the majority and minority groups as a fairness measure($\Delta_{SP}$).
	$$(TP_a + FP_a) = (TP_b + FP_b) $$
	$$\Delta_{SP}=(TP_a + FP_a)- (TP_b + FP_b) $$
	

	

    \item Equal opportunity is defined as the False negative rate (FNR) for all the sensitive attribute groups is equal, i.e., $P(-|+, a) = P(-|+,b) \ \ \forall a,b \in A$. We take the difference of FNR between the majority and minority groups to measure fairness($\Delta_{EO}$). 
	$$FNR  = \frac{FN_a}{TP_a+FN_a} = \frac{FN_b}{TP_b+FN_b}$$
	$$ \Delta_{EO}  = \frac{FN_a}{TP_a+FN_a} - \frac{FN_b}{TP_b+FN_b}$$
	
\end{itemize}

\subsection{Types of Missing Data } 
We consider missing data from two perspectives: the patterns of missingness and the attributes with missingness. 
The patterns of missingness in the data can occur in the following three ways~\cite{little2019statistical}.

\begin{itemize}
    \item {\bf Missing Completely at Random (MCAR):} The missing data is called MCAR if the missingness is independent of the observed and missing(unobserved) data.
    
    \item {\bf Missing at Random (MAR):} If the missingness is only dependent upon the observed data, the missing mechanism is MAR. The missing data is dependent on only the complete(observed) cases.
	
	\item {\bf Missing Not at Random (MNAR)}: For MNAR, the missingness is dependent on observed and unobserved data (missingness depends on complete and incomplete cases). In MNAR, the missingness occurs in the sensitive attribute depending on the value of the sensitive attribute.
\end{itemize}

We categorize missingness in attributes into two types. First, we consider missingness from sensitive attributes only (represented by $K_1$). Secondly, we consider missing from $K$ attributes, where $K$ is a predefined number. We select $K$ to be $10\%$ and $40\%$ of the total number of attributes ($K_{10}$, $K_{40}$) in the dataset.  We consider real graph datasets to assess fairness. The missing data is introduced artificially in the datasets, and $K$ most correlated attributes are selected. The sensitive attribute is also a part of $K_{10}$ and $K_{40}$ even if it is not the most correlated feature. The missing data are imputed using different imputation techniques, and fairness is computed. Given a sample $x_i=[s1,s2,....s_N]$, the probability that $s_j, \forall j \in \{1,2,...,N\}$ is missing is given in Table~\ref{tab:missing_types}. The $a$ and $b$ in MAR (Table~\ref{tab:missing_types})  represent the majority and the minority group, respectively. Note that producing missingness in sensitive attribute based on sensitive attribute will leads MAR to MNAR, due to which the first four rows of MAR are empty. Similarly, we do not include the sensitive attribute for $3$b, $4$b, $7$b, and $8$b as it will also lead  MAR to MNAR. Also note that all the probabilities in Table~\ref{tab:missing_types} are truncated within $[0,1]$. 

\begin{table}[t!]
\begin{center}
    
\resizebox{\columnwidth}{!}{%
\begin{tabular}{llllllll}
\toprule
K & MCAR & & MAR & & MNAR & \\
\midrule \midrule
 1 &  0.1   & (1a) & - & -  & 0.5 - $s_{j}$ & (1c)  \\
 1& 0.3 &  (2a)  &  -  & - &  0.5 - 0.2$s_{j}$ & (2c)\\
1&   0.5 & (3a) &  -    & - &  0.5 + 0.2$s_{j}$ & (3c)\\
 1 &0.7   &  (4a) &  -     & - & 0.5 +$s_{j}$  & (4c)\\
 10\% &0.1  &  (5a)  &   0.1+0.8$\times\mathbbm{1}_{a}$   &(1b) & 0.5 - $s_{j}$ &(5c)  \\
10\%  &   0.3 & (6a) &   0.1+0.8$\times\mathbbm{1}_{b}$   &(2b) &  0.5 - 0.2$s_{j}$  &(6c) \\
10\%  & 0.5  & (7a)  &   0.5 - 0.5$s_{K+j}$    & (3b) & 0.5 + 0.2$s_{j}$ & (7c) \\
10\%  &  0.7  & (8a) & 0.5 + 0.5$s_{K+j}$      & (4b) & 0.5 + $s_{j}$  & (8c)\\
40\%  & 0.1   & (9a) &  0.1+0.8$\times\mathbbm{1}_{a}$    &(5b) & 0.5 - $s_{j}$ & (9c)   \\
40\%  & 0.3   & (10a) &  0.1+0.8$\times\mathbbm{1}_{b}$    &(6b) &  0.5 - 0.2$s_{j}$ &(10c)  \\
40\%  & 0.5   & (11a) & 0.5 - 0.5$s_{K+j}$      &(7b) & 0.5 +0.2 $s_{j}$  &(11c) \\
40\%  & 0.7   &(12a)  &  0.5 + 0.5$s_{K+j}$     &(8b) & 0.5 + $s_{j}$ & (12c)\\
\bottomrule
\end{tabular}
}
\end{center}
\caption{Data missing mechanisms. Note that for MAR from sensitive attribute will transform it to MNAR. These are represented by '-' in MAR. Therefore, we have only considered missingness in other attributes based on sensitive attributes for MAR. $\mathbbm{1}_{a}$ and $\mathbbm{1}_{b}$ represents the indicator function with respect to $a$ and $b$ respectively.}
\label{tab:missing_types}
\end{table}

\subsection{Imputation Techniques}
We use four imputation techniques that are commonly used in data imputation research.

\paragraph*{\bf Mean imputation (Mean)}
The method imputes the missing with the mean of all the samples with observed values of the same attribute~\cite{spinelli2020missing}. 
It is easy to compute the mean, but it cannot capture the temporal trend in the data.

\paragraph*{\bf K-nearest neighbors (KNN)}
The method imputes the missing value using the KNNs that have observed values in the same attribute based on a distance metric to sample~\cite{spinelli2020missing}. KNN is a robust imputation technique and can capture complex data distribution easily. However, it is computationally expensive, and the results depend on the selection of hyperparameters.

\paragraph*{\bf Iterative SVD (SVD)}
The method imputes missing values based on matrix completion with iterative low-rank SVD decomposition inspired by~\cite{troyanskaya2001missing}. SVD can work for very sparse data. However, for large datasets, it is a computationally expensive operation.
\paragraph*{\bf Soft Imputation (SI)}
Matrix completion method by iterative soft thresholding of SVD decomposition, which is based on~\cite{you2020handling}. It uses the spectral regularization method for imputing incomplete matrices.

\section{Node Classification Algorithms}\label{sec:node class}
We use a range of algorithms, from \textit{fair oblivious} to \textit{fair node classifiers}.  Detail of these algorithms is given below. 

\subsection{Fairness Oblivious Node Classifiers}
Fair oblivious node classifiers do not consider or try to improve the fairness of the model. We select two fair oblivious classifiers, Node2Vec and Graph Convolutional Networks (GCN), since they are widely used by the research community in graph node classification.

\paragraph*{\textbf{Node2Vec}}
It is an algorithm to generate vector representations of nodes on a graph. The Node2Vec framework learns low-dimensional representations for nodes in a graph through the use of random walks. Besides reducing the engineering effort, representations learned by the algorithm lead to greater predictive power. 
The algorithm generalizes prior work, which is based on rigid notions of network neighborhoods, and argues that the added flexibility in exploring neighborhoods is the key to learning richer representations of nodes in graphs~\cite{grover2016node2vec,khosla2019comparative}.
\begin{figure}[t!]
\centering
\begin{subfigure}{.25\textwidth}
  \centering
  \includegraphics[scale=0.21] {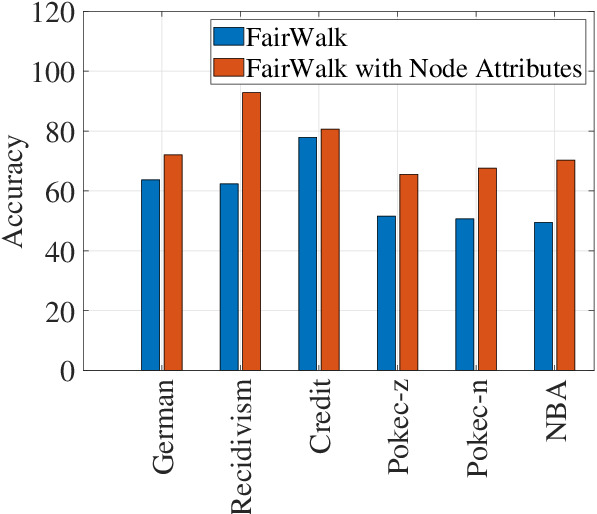}
  \caption{Accuracy}
  \label{fig:crosswalk_acc}
\end{subfigure}%
\begin{subfigure}{.25\textwidth}
  \centering
  \includegraphics[scale=0.21] {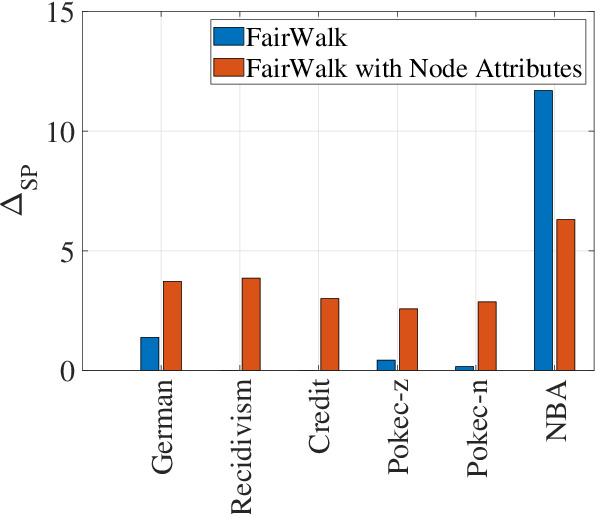}
  \caption{$\Delta_{SP}$}
  \label{fig:crosswalk_sp}
\end{subfigure}%
\\
\begin{subfigure}{.5\textwidth}
  \centering
  \includegraphics[scale=0.22] {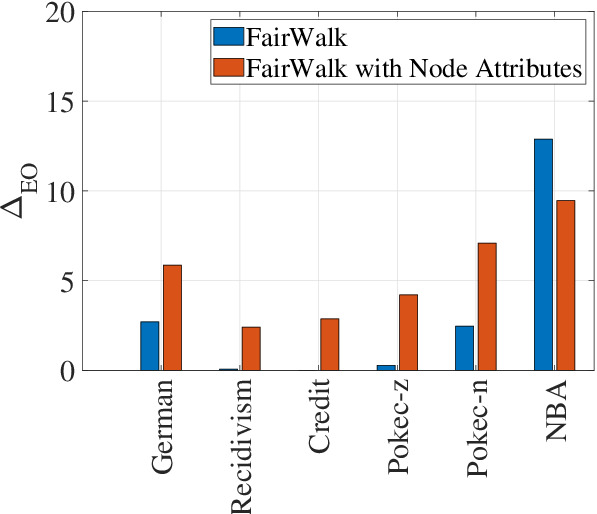}
  \caption{$\Delta_{EO}$}
  \label{fig:crosswalk_eo}
\end{subfigure}%
\caption{Comparison of CrossWalk with and without node attributes for all datasets. CrossWalk technique represents only node embeddings without introducing missingness in the data. German and Credit represent German Credit and Credit Defaulter datasets, respectively. Figure best seen in color.}
\label{fig:crosswalk}
\end{figure}

\paragraph*{\textbf{Graph Convolutional Networks (GCN)}}
This method is similar to convolutional neural networks (CNNs) in terms of weight sharing. The main difference lies in the data structure, where GCNs are the generalized version of CNNs that can work on data with underlying non-regular structures. Providing information about edges to GCNs enables the model to learn the features of neighboring nodes. This mechanism can be seen as a message-passing operation along the nodes within the graph~\cite{kipf2016semi}. After hyperparameter tuning, we select one convolution layer of size $8\times8$ with RELU activation function followed by a dense layer with a sigmoid activation function. This architecture is used for all datasets.

\subsection{Fair Node Classifiers}
The goal of fair node classifiers is to improve the fairness of the model without compromising accuracy. They must achieve this by modifying the classifier's random walk process or objective function.

\paragraph*{\textbf{FairWalk}~\cite{Rahman2019FairwalkTF}}
This method relies on a modification of the random walks to induce fairness in node embeddings. This method modifies the transition probability of Node2Vec for the generation of unbiased traces. It works well on graphs in which the neighboring nodes have different sensitive attributes.
    
\paragraph*{\textbf{CrossWalk}~\cite{khajehnejad2022crosswalk}}
This method enhances the fairness of various graph algorithms applied to node embeddings, including influence maximization, link prediction, and node classification. It is based on a re-weighting procedure for building the random walks and can therefore be used with any random walk-based algorithms, including Node2Vec. CrossWalk aims to address the shortcomings of FairWalk by assigning more weights to both edges that connect nodes on the boundary of the same groups and edges connecting nodes from different groups.

\section{Experimental Setup}\label{sec:experimental_setup}
In this section, we describe the datasets used in our study of fairness. We also provide implementation details such as pre-processing of data and experimental parameters.

\subsection{Dataset Statistics}
For this study, we used six datasets, ranging from social network graphs to credit defaulter and bail datasets. The statistics of these datasets are given in Table~\ref{tabl_benchmark}. The details are given below.

\subsubsection{Pokec}
It is a social network in Slovakia~\cite{dai2021say}, similar to Facebook and Twitter. Nodes are user profiles, and edges are friendships between the nodes. 
This dataset contains anonymized data from the whole social network in $2012$. User profiles contain gender, age, hobbies, interests, education,
working field and etc. The original Pokec dataset contains millions
of users. Based on the provinces that users belong to, we sampled
two datasets named Pokec-z and Pokec-n. The Pokec-z and
Pokec-n consists of users belonging to two major regions of the corresponding
provinces. We treat the region as a sensitive attribute.
The classification task is to predict the working field of the users.

\begin{figure}[t!]
\centering
\begin{subfigure}{.25\textwidth}
  \centering
  \includegraphics[scale=0.21] {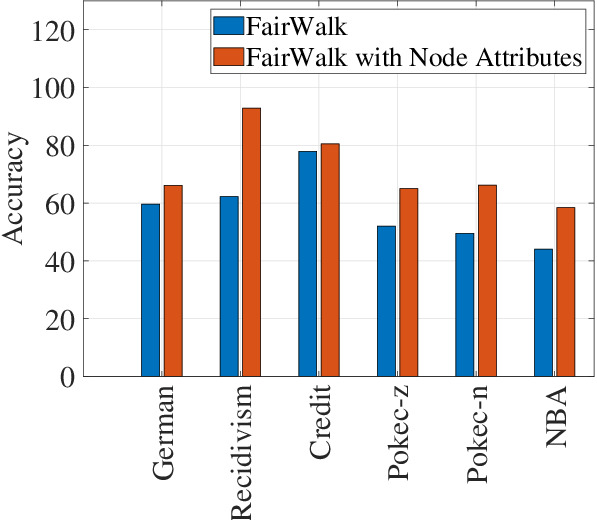}
  \caption{Accuracy}
  \label{fig:fairwalk_acc}
\end{subfigure}%
\begin{subfigure}{.25\textwidth}
  \centering
  \includegraphics[scale=0.21] {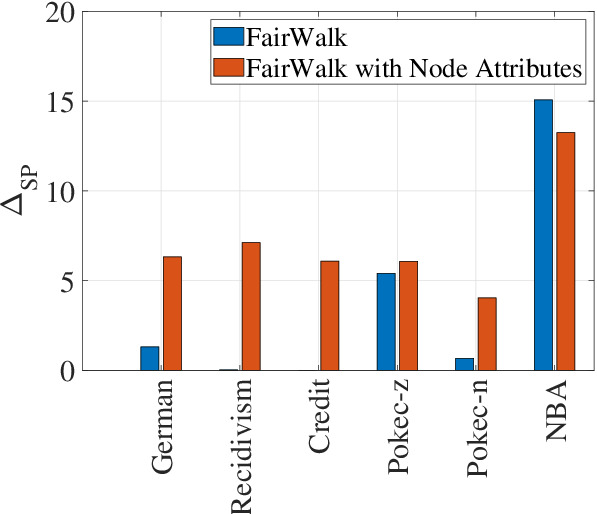}
  \caption{$\Delta_{SP}$}
  \label{fig:fairwalk_sp}
\end{subfigure}%
\\
\begin{subfigure}{.5\textwidth}
  \centering
  \includegraphics[scale=0.22] {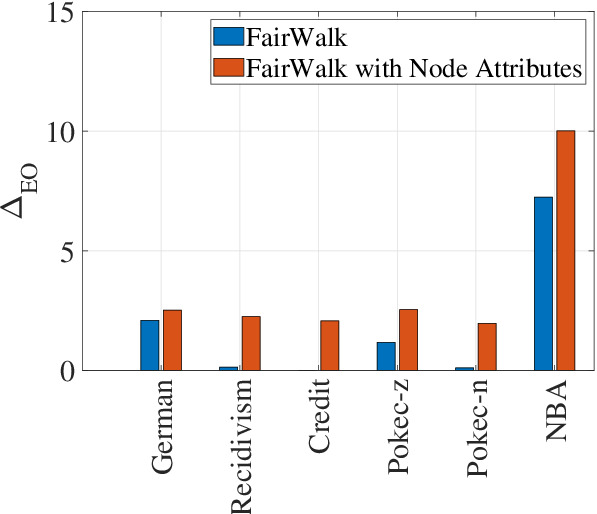}
  \caption{$\Delta_{EO}$}
  \label{fig:fairwalk_eo}
\end{subfigure}%
\caption{Comparison of FairWalk with and without node attributes for all datasets. FairWalk technique represents only node embeddings without introducing missingness in the data. German and Credit represent German Credit and Credit Defaulter datasets, respectively. Figure best seen in color.}
\label{fig:fairwalk}
\end{figure}
\subsubsection{NBA}
It contains performance statistics of players in the 2016-2017~\cite{dai2021say} season along with information such as nationality, age, and salary for $400$ NBA basketball players. Nodes are players, and edges are their relationship on Twitter. 
To obtain the graph that links the NBA players together, we collect the relationships of the NBA basketball players on Twitter with its official crawling API. We binarize the nationality into two categories, i.e., U.S. players and overseas players, which are used as sensitive attributes. The classification task is to predict whether or not the player's salary is over the median.

\subsubsection{German Credit Graph} 
The German credit graph represents clients in a German bank that are connected based on the similarity of their credit accounts. The task is to classify clients into good vs. bad credit risks considering the client's gender as a sensitive attribute. The German Graph credit dataset~\cite{agarwal2021towards,dong2022edits,zhang2021multi} classifies people described by a set of attributes as good or bad credit risks. It consists of attributes like gender, loan amount, and other account-related features of $1,000$ clients(nodes) and $25000$ edges. The task is to classify clients into good vs. bad credit risks considering the client's gender as a sensitive attribute. Other attributes include age, single, loan information, employment, customer finance, and assets data.

\subsubsection{Credit Defaulter Graph }
The Credit defaulter graph has $30000$ nodes representing individuals that we connected based on the similarity of their spending and payment patterns~\cite{agarwal2021towards,dong2022edits,fan2021fair,zhang2021multi}. The task is to predict whether an individual will default on the credit card payment or not while considering age as a sensitive attribute. It contains features like education, credit history, age, and features derived from their spending and payment patterns.
\subsubsection{Recidivism }
The Recidivism graph has $18876$ nodes representing defendants who got released on bail at the U.S state courts during $1990$-$2009$~\cite{agarwal2021towards,dong2022edits,fan2021fair,zhang2021multi}. Defendants are connected based on the similarity of past criminal records and demographics. The goal is to classify defendants into bail (i.e., unlikely to commit a violent crime if released) vs. no bail (i.e., likely to commit a violent crime), considering race information as the
protected attribute. The data contains education, age, and other personal attributes.

\begin{table}[t!]
	\centering
	\resizebox{0.5\textwidth}{!}{%
		\begin{tabular}{lccp{1cm}p{1cm}c}
			\toprule
			\textbf{Dataset}      &  $\mathbf{|V_G|}$ & $\mathbf{|E_G|}$ & \textbf{Average Degree} & \textbf{Density}  &  \textbf{$\vert$Attributes$\vert$} \\ 
			\midrule \midrule
			Pokec-n   & 66569 & 517047 & 4.71 &  0.00071 &   268 \\ [.04in]
			Pokec-z   & 67796 & 617958 & 5.17 & 0.00064 &   279 \\ [.04in]
			NBA   & 400 & 10621 & 45.90 & 0.148 &   98 \\ [.04in]
			German Credit   & 1000 & 21742 & 43.48 &  0.043&  30 \\ [.04in]
			Recidivism   & 18876  & 311870 & 33.04 & 0.0017 & 19  \\ [.04in]
			Credit Defaulter   & 30000 & 1421858  & 94.79 & 0.0031 &  15 \\ [.04in]
			\bottomrule
		\end{tabular}
	}
	\caption{Datasets statistics, where $\mathbf{|V_G|}$ and $\mathbf{|E_G|}$ represent the number of nodes and edges in the graph, respectively.}
	\label{tabl_benchmark}
\end{table}

\subsection{Implementation Details}
The NBA dataset contains missing values. We remove the nodes with missing values, after which the size of the dataset is around $300$ nodes. We also remove the corresponding edges of missing nodes. Similarly, Pokec-z and Pokec-n also contain a large number of nodes with missing attributes, which we remove, after which the number of nodes for Pokec-z and Pokec-n is around $6500$ and $8000$, respectively. 

We first transform the datasets with Min-Max normalization. We use $70\%$ of the data for training and validation purposes. The remaining $30\%$ data is used for testing purposes. We stratify the train test split with respect to sensitive attributes to counter any imbalance. The missingness is artificially introduced in the training data according to Table~\ref{tab:missing_types}, the missing data is imputed, node classifiers are trained using imputed data, and results are computed. Experiments are performed on core i7, a $9^{th}$ generation machine with $16$ GB memory. All experiments are performed five times, and the mean results are reported.
\begin{figure}[t!]
\centering
  \includegraphics[width=\linewidth] {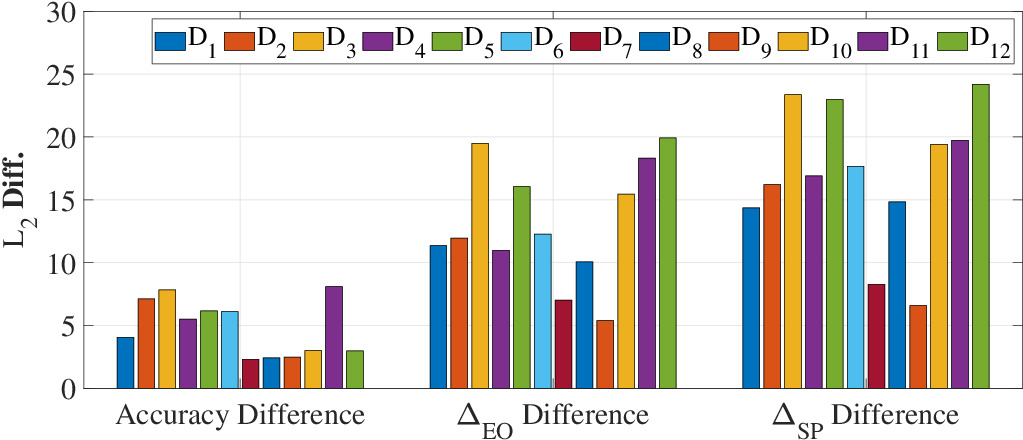}
  \caption{Results comparison using different imputation techniques for German Credit dataset and MCAR. Y-axis represents the difference of results in terms of $L_{2}$ norm. $D_{1}$ represents the difference in results from KNN: KNN-Mean CrossWalk, $D_{2}$:KNN-SVD CrossWalk, $D_{3}$:KNN-SI CrossWalk, $D_{4}$:KNN-Mean FairWalk, $D_{5}$:KNN-SVD FairWalk, $D_{6}$:KNN-SI FairWalk, $D_{7}$:KNN-Mean GCN, $D_{8}$:KNN-SVD GCN, $D_{9}$:KNN-SI GCN, $D_{10}$:KNN-Mean Node2Vec, $D_{11}$:KNN-SVD Node2Vec, $D_{12}$:KNN-SI Node2Vec. Figure best seen in color.}
  \label{fig:MCAR_german_impu_diff}
 \end{figure}
For the implementation of FairWalk\footnote[1]{\url{https://github.com/urielsinger/fairwalk}} and Crosswalk\footnote[2]{\url{https://github.com/ahmadkhajehnejad/CrossWalk}} we use their respective GitHub repository. For the implementation of Node2Vec and GCN we use Stellargrap\footnote[3]{\url{https://github.com/stellargraph/stellargraph}} library. 
We set the embedding dimension as $128$ for CrossWalk, FairWalk, and Node2Vec. The remaining parameters are the same as stated in their papers ~\cite{grover2016node2vec,khajehnejad2022crosswalk,Rahman2019FairwalkTF}. Since node embeddings only use edge information, they do not depend upon node features. There is no point in missing data imputation in CrossWalk, FairWalk, and Node2Vec. To counter this issue, we concatenate node embeddings with respective node features. Logistic regression is used as a classifier in Node2vec, CrossWalk, and FairWalk. The architecture of GCN is already explained in Section~\ref{sec:node class}. Binary cross entropy (BCE) is used as a loss function for GCN. An ADAM optimizer with a $0.01$ learning rate is used to train GCN. For the implementation of Mean, KNN, SVD, and SI, we have used an open source library FancyImpute\footnote[4]{\url{https://github.com/iskandr/fancyimpute}}. The hyperparameters of KNN, SVD, and SI are the same as those given by FancyImpute. All our code and datasets used are available at Github\footnote[5]{\url{https://github.com/harisalimansoor/FairnessGraphDataImputation}}.

\section{Results and Discussions}\label{sec:results}
\begin{figure}[t!]
\centering
\begin{subfigure}{\linewidth}
  \centering
  \includegraphics[width=\linewidth] {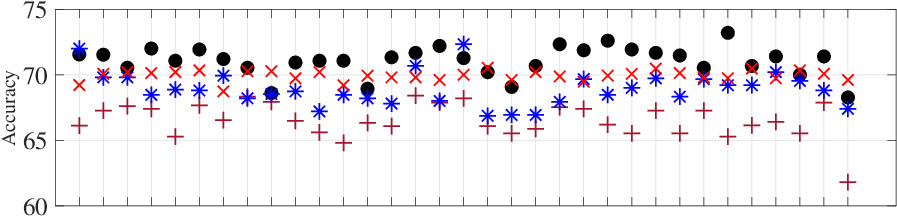}
  \caption{Accuracy}
  \label{fig:ger_acc}
\end{subfigure}
\begin{subfigure}{\linewidth}
  \centering
  \includegraphics[width=\linewidth] {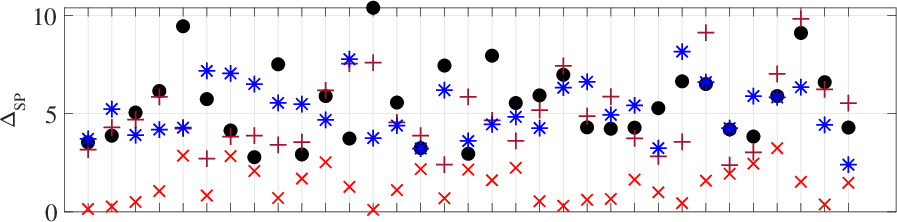}
  \caption{$\Delta_{SP}$}
  \label{fig:ger_sp}
\end{subfigure}
\begin{subfigure}{\linewidth}
  \centering
  \includegraphics[width=\linewidth] {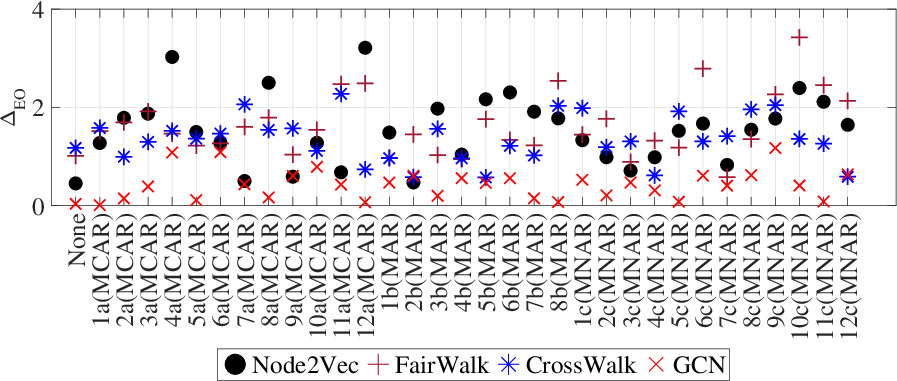}
  \caption{$\Delta_{EO}$}
  \label{fig:ger_eo}
\end{subfigure}
\caption{Results comparison of different classifiers and missing mechanisms for German Credit dataset. None represents data without missingness, remaining scenarios are explained in Table~\ref{tab:missing_types}. Figure best seen in color.}
\label{fig:ger scatter}
\end{figure}

\begin{figure}[t!]
\centering
\begin{subfigure}{\linewidth}
  \centering
  \includegraphics[width=\linewidth] {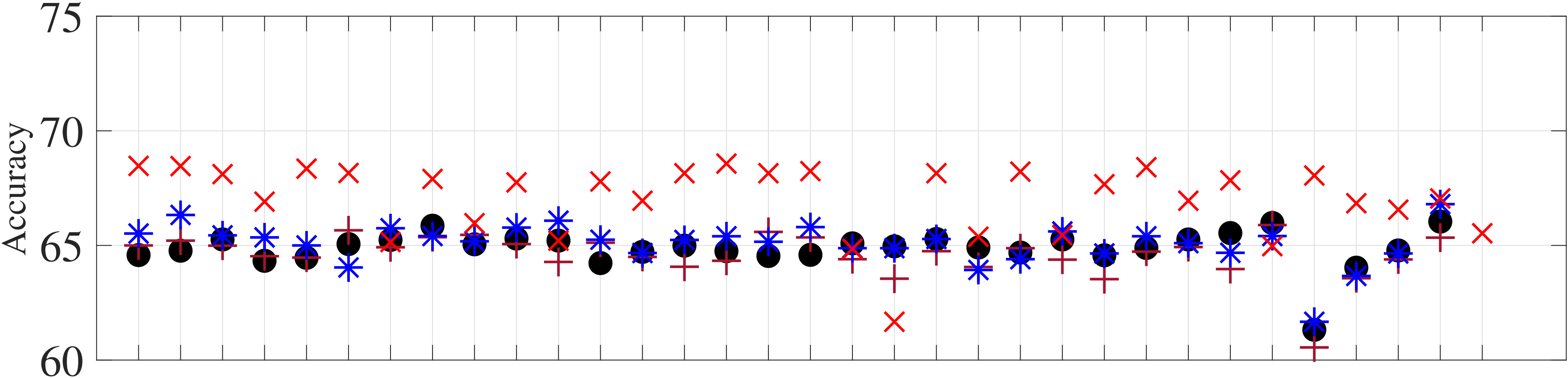}
  \caption{Accuracy}
  \label{fig:pokn_acc}
\end{subfigure}
\begin{subfigure}{\linewidth}
  \centering
  \includegraphics[width=\linewidth] {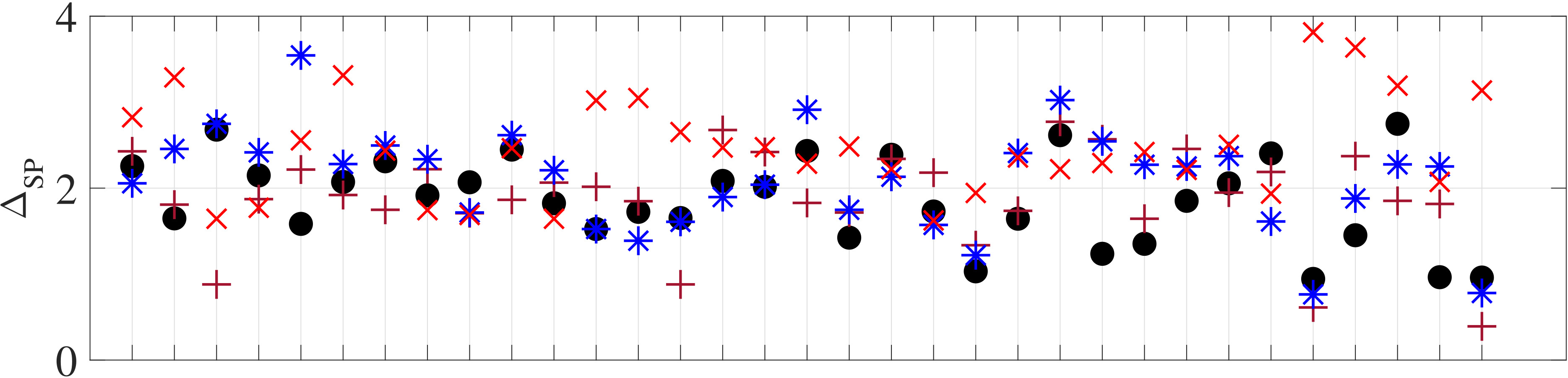}
  \caption{$\Delta_{SP}$}
  \label{fig:pokn_sp}
\end{subfigure}
\begin{subfigure}{\linewidth}
  \centering
  \includegraphics[scale=0.28,angle =90] {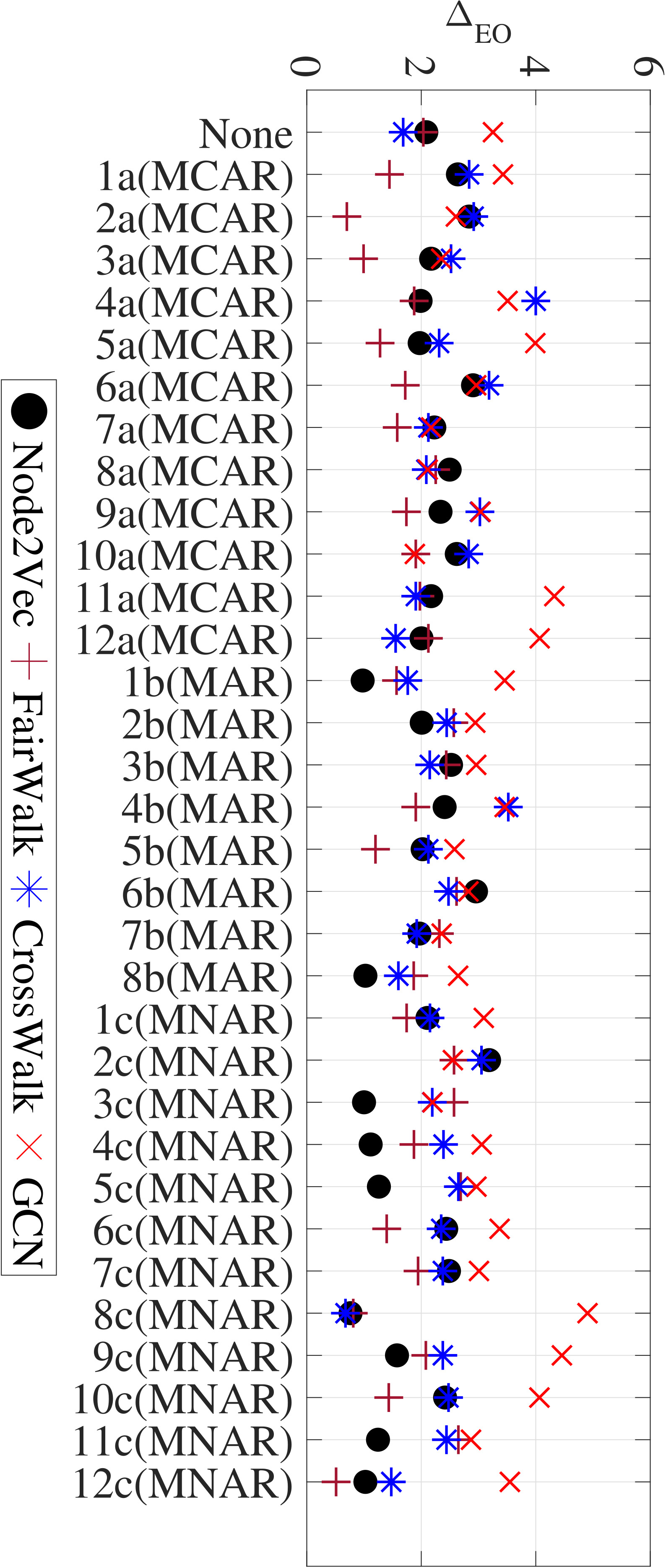}
  \caption{$\Delta_{EO}$}
  \label{fig:pokn_eo}
\end{subfigure}
\caption{Results comparison of different classifiers and missing mechanisms for the Pokec-n dataset. None represents data without missingness, remaining scenarios are explained in Table~\ref{tab:missing_types}. Figure best seen in color.}
\label{fig:Pokec-n scatter}
\end{figure}

In this section, we present our findings regarding the impact of missing data imputation on the performance and fairness of graph node classifiers. We use statistical parity and equal opportunity as measures of fairness and accuracy to assess the classifiers' performance. Specifically, we investigate the effect of (i) imputation techniques on fairness or accuracy, (ii) an increase in the probability of missingness, (iii) an increase in the number of attributes of missingness,  and (iv) a fair oblivious versus fair classifier. Due to the page limit, we only present results for KNN imputation. Our main findings are reported in the form of a scatter plot in Figure~\ref{fig:ger scatter} (German Credit),
Figure~\ref{fig:Pokec-n scatter} (Pokec-N),
Figure~\ref{fig:Pokec-z scatter} (Pokec-Z),
Figure~\ref{fig:credit defaulter scatter} (Credit Defaulter),
Figure~\ref{fig:recidivism scatter} (Recidivism), and Figure~\ref{fig:NBA scatter} (NBA).

\subsection{Effect of MCAR}
In this section, we show the effect of increasing values of missing probability and increasing value of $K$.
\subsubsection{Impact of Increasing Missing Probability}
We find that as the missing probability increases, both $\Delta_{SP}$ and $\Delta_{EO}$ tend to increase for almost all datasets and all classifiers. This can be observed from $1$a (MCAR) to $4$a (MCAR) in Figure~\ref{fig:ger scatter},
~\ref{fig:Pokec-n scatter},
~\ref{fig:Pokec-z scatter},
~\ref{fig:credit defaulter scatter},
~\ref{fig:recidivism scatter}, and ~\ref{fig:NBA scatter}. The effect on German Credit dataset is quite worse and $\Delta_{SP}$ increases from $7.08$ (1a) to $18.92$ (4a) which is more than $150 \%$ increase in $\Delta_{SP}$ (Figure~\ref{fig:ger_sp}). In the case of equal opportunity, the German dataset also performs worse with $\Delta_{EO}$  increases from $2.25$ (1a) to $15.12$ (4a), which is more than $500 \%$ increase. The Credit Defaulter dataset also shows a similar behaviour (Figure~\ref{fig:credit defaulter scatter}), $\Delta_{SP}$ increases from $4.36$ (1a) to $8.77$ (4a) which is about $100 \%$ increase. Similarly, the equal opportunity $\Delta_{EO}$ increases from $2.84$ (1a) to $5.79$ (4a) which is also more than $100 \%$ increase. Recidivism, Pokec-z, Pokec-n, and NBA datasets also produce negative results on fairness as missing probability increases, but they are not as severe as the German and Credit Defaulter dataset. Another interesting observation is that equal opportunity difference, although increases with missing probability but it remains more stable than parity difference. Increasing the missing probability also negatively affects accuracy but is not as adverse as fairness. German credit, NBA, and Pokec-n show a decrease in accuracy in the case of 4a(MCAR) for different classification algorithms  (Figure~\ref{fig:ger scatter},\ref{fig:Pokec-n scatter},\ref{fig:NBA scatter}).

\subsubsection{Impact Of Increasing K}
\begin{figure}[t!]
\centering
\begin{subfigure}{\linewidth}
  \centering
  \includegraphics[width=\linewidth] {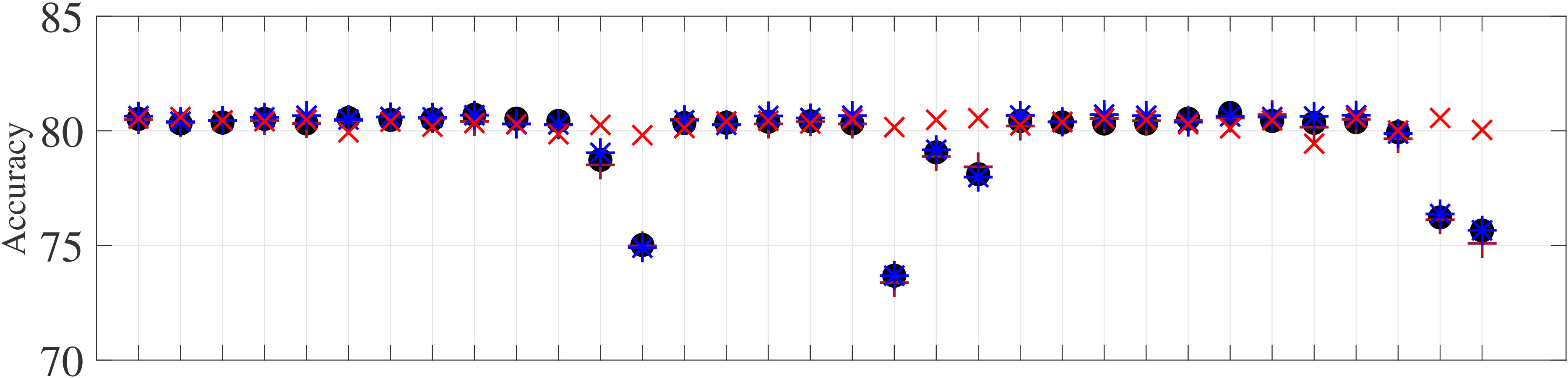}
  \caption{Accuracy}
  \label{fig:cre_acc}
\end{subfigure}
\begin{subfigure}{\linewidth}
  \centering
  \includegraphics[width=\linewidth] {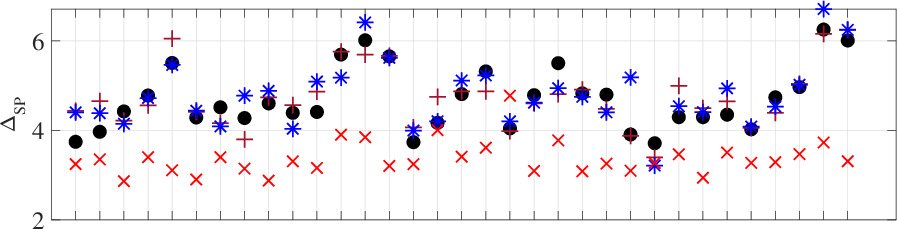}
  \caption{$\Delta_{SP}$}
  \label{fig:cre_sp}
\end{subfigure}
\begin{subfigure}{\linewidth}
  \centering
  \includegraphics[width=\linewidth] {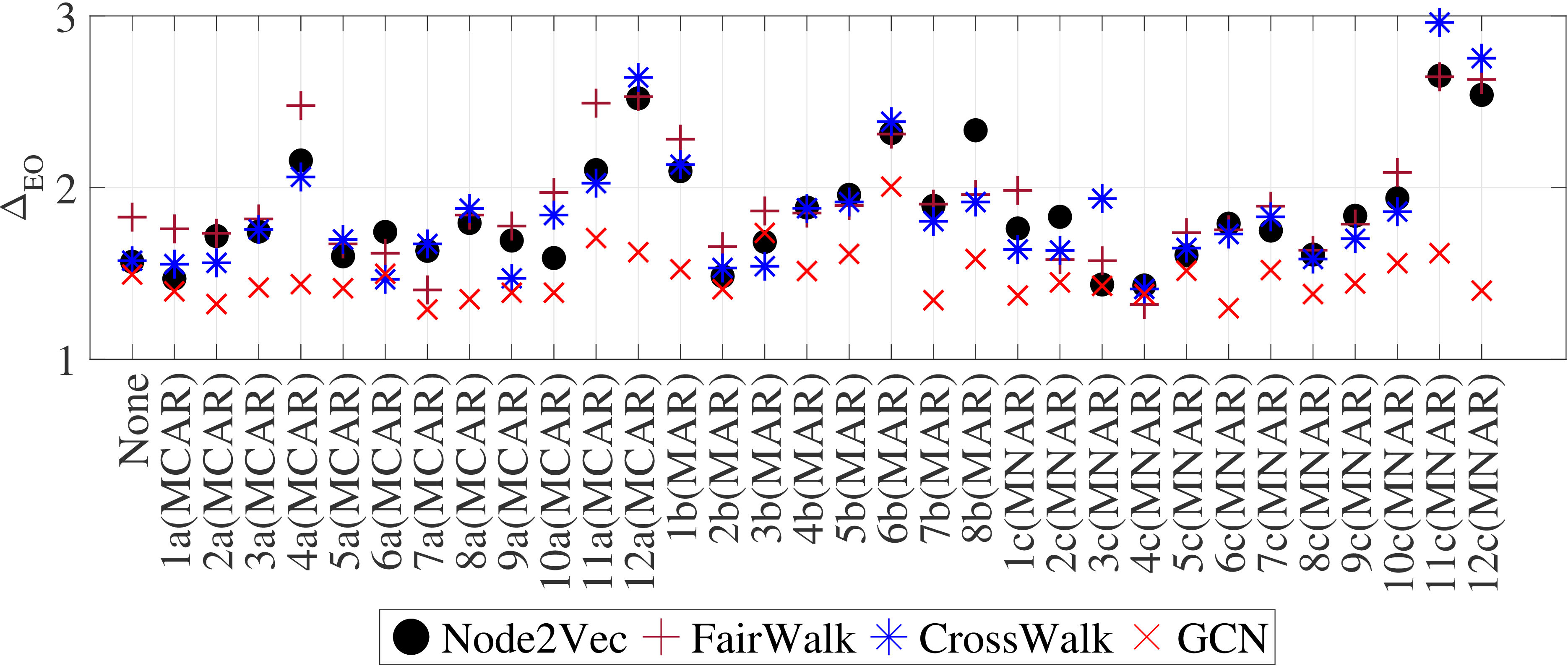}
  \caption{$\Delta_{EO}$}
  \label{fig:cre_eo}
\end{subfigure}
\caption{Results comparison of different classifiers and missing mechanisms for Credit Defaulter dataset. None represents data without missingness, remaining scenarios are explained in Table~\ref{tab:missing_types}. Figure best seen in color.}
\label{fig:credit defaulter scatter}
\end{figure}

As the number of attributes of missingness $K$ increases, a general trend of increase in both $\Delta_{SP}$ and $\Delta_{EO}$ is observed for almost all datasets and all methods. This trend is present in almost all probabilities ($0.1,0.3,0.5,0.7$), keeping probability constant and changing $K$ from $10\%$ to $40\%$ increases the  $\Delta_{SP}$ and $\Delta_{EO}$.
We can see this by comparing $1a-4a$, $5a-8a$, and $9a-12a$ in the result figures. The accuracy is also more severely affected by increasing $K$, which can be observed from $8$a and $12$a in the results tables. The accuracy remains consistent for the Credit Defaulter dataset showing robustness(~\ref{fig:credit defaulter scatter}).
We have noticed some anomalies in the NBA dataset, which can be explained by the small sample size ($300$). Due to the small sample size, little changes in the confusion matrix cause the fairness and accuracy to fluctuate (Figure~\ref{fig:NBA scatter}).
By analyzing the results, we conclude that increasing the probability and $K$ generally have a negative impact on fairness and accuracy. However, the specific effect varies greatly on the dataset and the algorithm used.

\subsection{Effect of MAR }
In MAR, 1b and 5b represent missingness based on the majority attribute, while 2b and 6b represent MAR from minority attributes.
We can observe that both fairness measures are more affected in 5b and 6b as compared to 1b and 2b, concluding increasing $K$ causes more bias in the case of MAR on the basis of the sensitive attribute.

\begin{figure}[t!]
\centering
\begin{subfigure}{\linewidth}
  \centering
  \includegraphics[width=\linewidth] {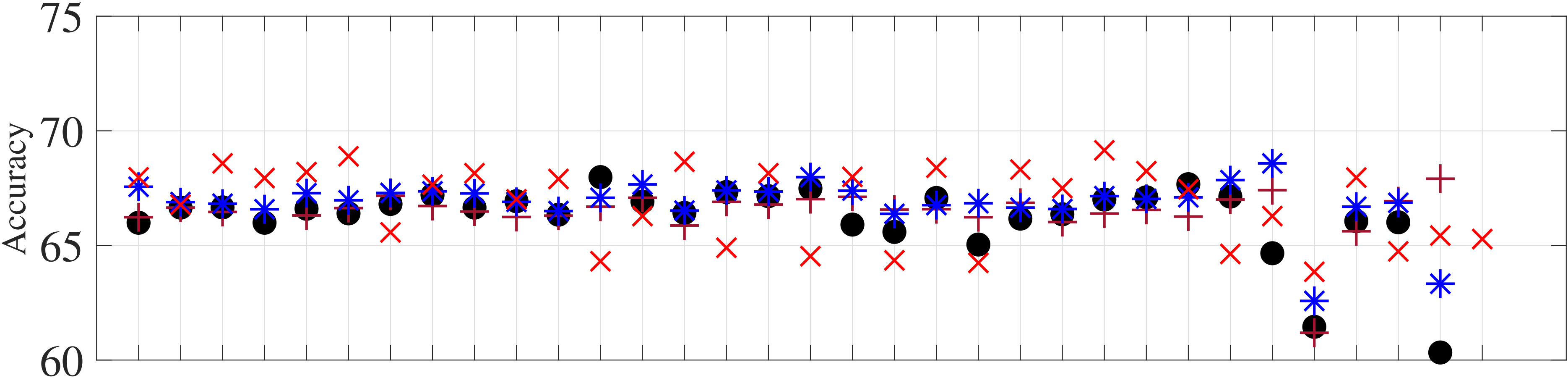}
  \caption{Accuracy}
  \label{fig:pokz_acc}
\end{subfigure}
\begin{subfigure}{\linewidth}
  \centering
  \includegraphics[width=\linewidth] {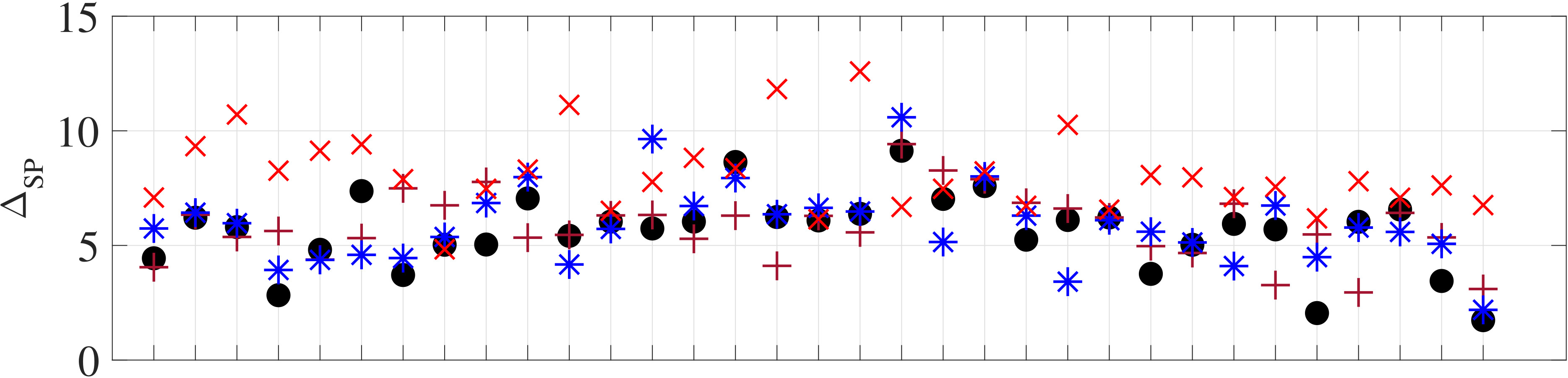}
  \caption{$\Delta_{SP}$}
  \label{fig:pokz_sp}
\end{subfigure}
\begin{subfigure}{\linewidth}
  \centering
  \includegraphics[scale=0.28,angle =90] {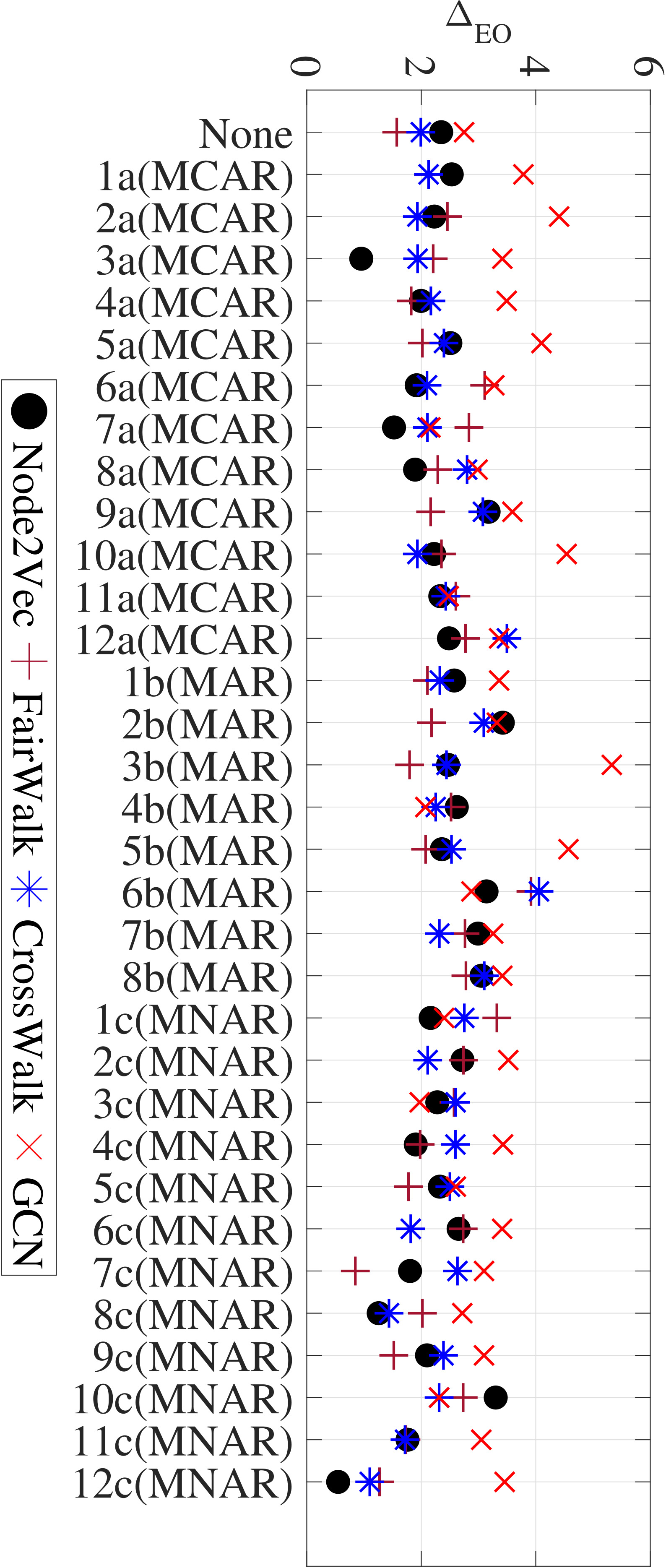}
  \caption{$\Delta_{EO}$}
  \label{fig:pokz_eo}
\end{subfigure}
\caption{Results comparison of different classifiers and missing mechanisms for the Pokec-z dataset. None represents data without missingness, remaining scenarios are explained in Table~\ref{tab:missing_types}. Figure best seen in color.}
\label{fig:Pokec-z scatter}
\end{figure}


In some datasets (German Credit, Credit Defaulter), 1b is more affected in terms of fairness as compared to 2b, while in the remaining datasets, we can observe the opposite behavior. This is due to the sensitive class imbalance. In German Credit and Credit Defaulter datasets, the samples with majority sensitive attributes are more than the minority; thus, MAR using majority attributes produces more imbalance. Another observation from comparing 3b, 4b to 1b, 2b (5b, 6b to 7b, 8b) is that MAR on the basis of sensitive attributes affects the results more as compared to MAR based on other attributes (3b, 4b). This also depicts the effect of sensitive attributes on fairness.
In MAR, we see an effect of fairness accuracy trade-off in some cases, where a decrease in bias (increase in fairness) is associated with a decrease in accuracy. This behavior is also reported by~\cite{Fairness_in_missing_data}. We can see this behavior in 7b and 6b for the German credit dataset, 6b and 8b in Recidivism dataset, 6b in the credit defaulter dataset, 8b in the Pokec dataset, and 5b and 8b in the NBA dataset.

\subsection{Effect of MNAR}
We assume that node attributes follow a uniform distribution. Under this assumption, the probability of missingness increases from 1c to 4c and similarly from 5c to 8c (9c to 12c). Moreover, 1c, 5c, and 9c have the same probability of missing. Only the number of attributes of missingness increases. 
We notice that as the missing probability increases, both fairness measures tend to increase for almost all datasets and all classifiers for MNAR.
This can be observed from $1c$ to $4c$ in the result figures. This behavior can also be observed when we increase K in 5c to 8c (9c to 12c).
We can also observe that as the number of attributes of missingness $K$ increases, an increase in both $\Delta_{SP}$ and $\Delta_{EO}$ is observed for almost all datasets and all methods. This is particularly observed in 11c and 12c in almost all datasets and methods. In MNAR, we also observe an effect of fairness accuracy balance in some cases, where a decrease in bias (increase in fairness) is associated with a decrease in accuracy, as reported in MAR too. We can notice that such kind of fairness accuracy trade-off widely exists in all datasets under MNAR.

\subsection{Effect of Imputation Techniques}
\begin{figure}[t!]
\centering
\begin{subfigure}{\linewidth}
  \centering
  \includegraphics[width=\linewidth] {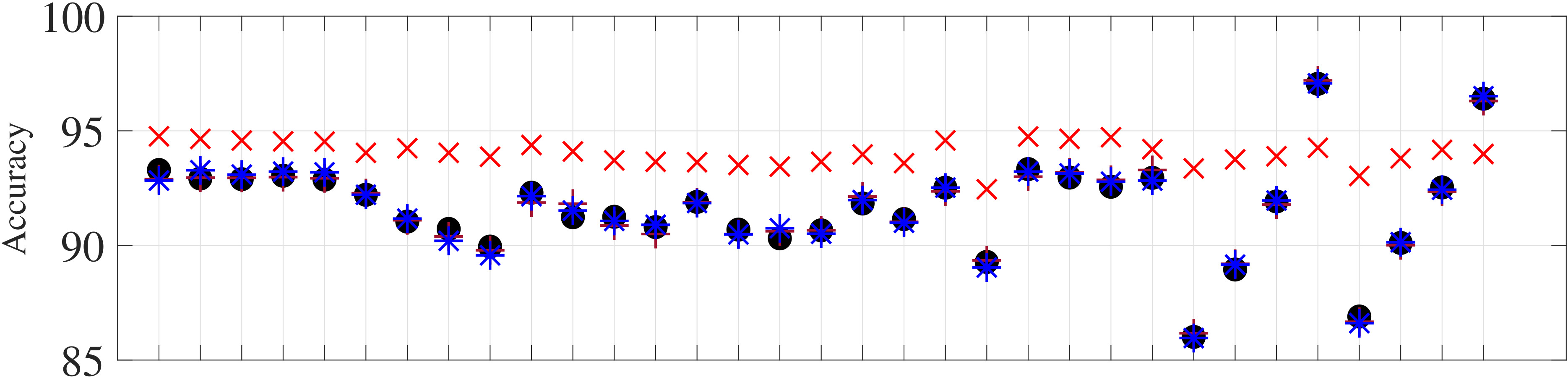}
  \caption{Accuracy}
  \label{fig:rec_acc}
\end{subfigure}
\begin{subfigure}{\linewidth}
  \centering
  \includegraphics[width=\linewidth] {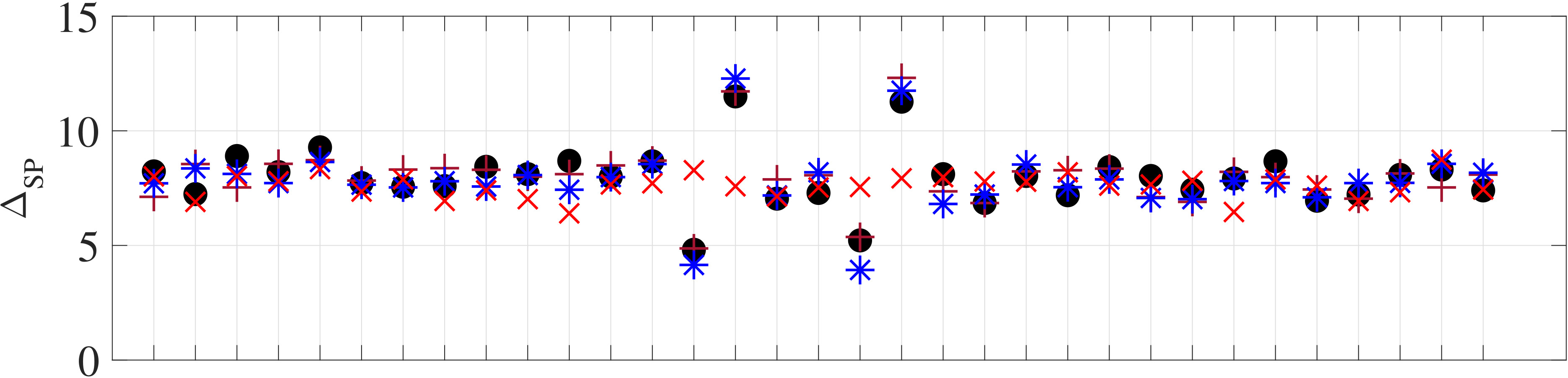}
  \caption{$\Delta_{SP}$}
  \label{fig:rec_sp}
\end{subfigure}
\begin{subfigure}{\linewidth}
  \centering
  \includegraphics[scale=0.28,angle =90] {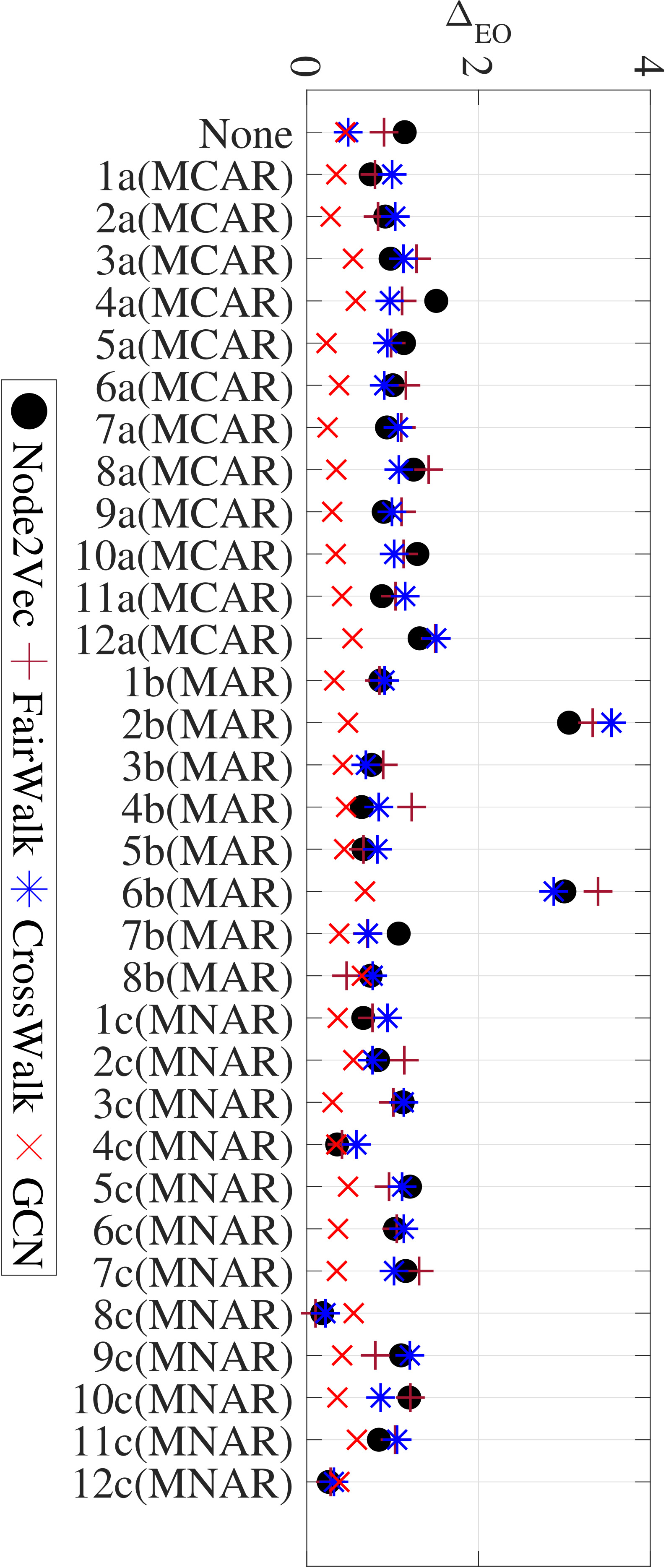}
  \caption{$\Delta_{EO}$}
  \label{fig:rec_eo}
\end{subfigure}
\caption{Results comparison of different classifiers and missing mechanisms, for Recidivism dataset. None represents data without missingness, remaining scenarios are explained in Table~\ref{tab:missing_types}. Figure best seen in color.}
\label{fig:recidivism scatter}
\end{figure}

\begin{figure}[t!]
\centering
\begin{subfigure}{\linewidth}
  \centering
  \includegraphics[width=\linewidth] {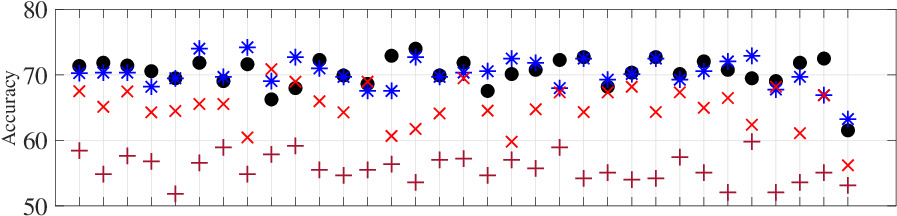}
  \caption{Accuracy}
  \label{fig:nba_acc}
\end{subfigure}
\begin{subfigure}{\linewidth}
  \centering
  \includegraphics[width=\linewidth] {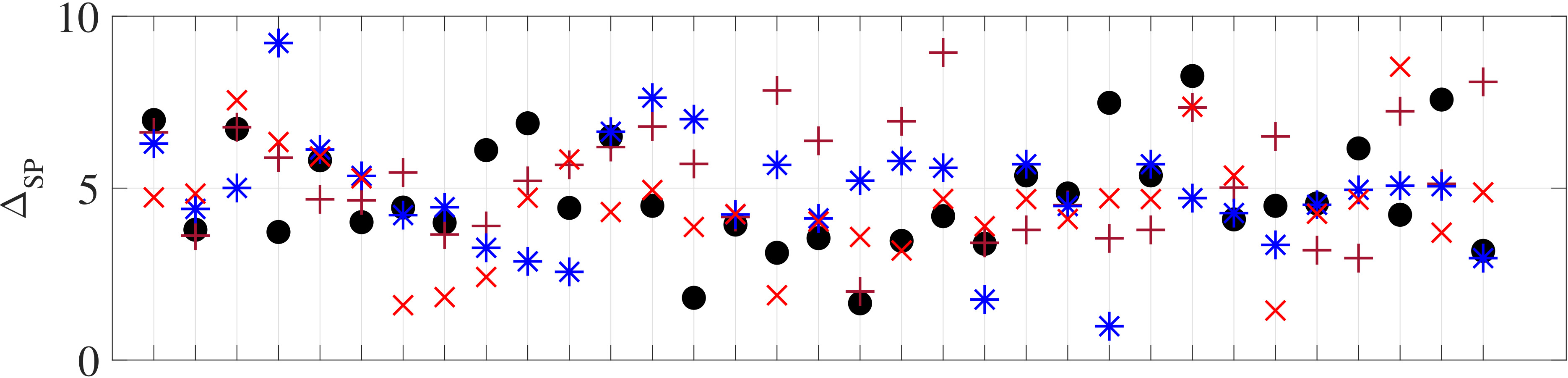}
  \caption{$\Delta_{SP}$}
  \label{fig:nba_sp}
\end{subfigure}
\begin{subfigure}{\linewidth}
  \centering
  \includegraphics[width=\linewidth] {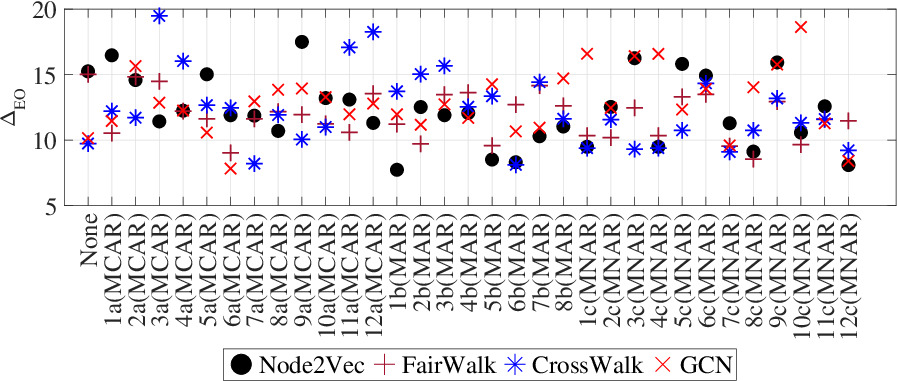}
  \caption{$\Delta_{EO}$}
  \label{fig:nba_eo}
\end{subfigure}
\caption{Results comparison of different classifiers and missing mechanisms for NBA dataset. None represents data without missingness, remaining scenarios are explained in Table~\ref{tab:missing_types}. Figure best seen in color.}
\label{fig:NBA scatter}
\end{figure}

We use four imputation techniques to validate our experiments. From the results, we found that although imputation techniques affect fairness and accuracy, they are not that prominent. This is why we only show results of KNN in Figure~\ref{fig:ger scatter},
~\ref{fig:Pokec-n scatter},
~\ref{fig:Pokec-z scatter},
~\ref{fig:credit defaulter scatter},
~\ref{fig:recidivism scatter}, and ~\ref{fig:NBA scatter}. To compare the different imputation methods and their results, we compute the  $L_2$ norm difference of fairness and accuracy results of different imputation techniques(Mean, SVD, SI) from KNN. Due to the page limit, we have shown results of only MCAR for the German Credit dataset in Figure~\ref{fig:MCAR_german_impu_diff}. We use KNN as a baseline for this comparison. Figure~\ref{fig:MCAR_german_impu_diff} provides valuable insights into the comparison of different imputation methods and classification techniques. Among all methods, GCN is closest to each other in terms of the classifier with the lowest difference in results. This implies that GCN is least affected by any imputation method. We can observe this in both accuracy and fairness measures ($D_7, D_8, D_9$). Similarly, Node2Vec and CrossWalk have the worst results and are badly affected by the imputation method. In terms of the magnitude of difference, accuracy is least affected by the selection of a specific imputation method, while parity difference is most affected with $L_2$ norm difference as high as 25 ($D_{10}, D_{11}, D_{12}$).

\subsection{Comparison of Fair Oblivious vs. Fair Classifiers}
We use two fair oblivious classifiers (Node2Vec, GCN) and two fair classifiers (FairWalk, CrossWalk). As stated earlier, we concatenate node embeddings with node features for Node2Vec, FairWalk, and CrossWalk. It is worth noting that using FairWalk and CrossWalk with only node embedding does improve fairness, but at the cost of compromised accuracy Figure~\ref{fig:fairwalk},~\ref{fig:crosswalk}. 
Another interesting observation comes from NBA data. Instead of improving fairness, Both FairWalk and CrossWalk reduce fairness, increasing bias. This can be explained in terms of the very small sample size of the NBA dataset. 

Concatenating node attributes increases accuracy, but fairness is compromised. Also, the increase in accuracy is not as compared to Node2Vec or GCN. This implies that the notion of fairness in FairWalk and CrossWalk is restricted to edges, and a true fair node classifier is still an open task. A trade-off between fairness and accuracy still exists in FairWalk and CrossWalk. We have explained earlier that such a trade-off further increases in MAR and MNAR.
Another interesting observation comes from GCN (fair oblivious classifier), which performs best in terms of fairness without compromising accuracy. For the Recidivism dataset, GCN also performs best in terms of accuracy Figure~\ref{fig:recidivism scatter}. This is due to the robust design of GCN, which integrates both edges and features. Incorporating fairness in GCN can further improve fairness without compromising accuracy.

\section{Conclusion}\label{sec:conclusions}
In this work, we explore the effect of missing data imputation on the fairness of graph node classifiers. We performed a detailed experimental study using six datasets, different imputation techniques, fair oblivious, and fair node classifiers. Our results demonstrate that severe fairness issues exist in missing data imputation of graphs, describing the first known empirical research in this direction. Although fair node classifiers improve fairness, it comes at the cost of compromised accuracy. Moreover, fair node classifiers are also affected by data imputation. A similar trade-off between accuracy and fairness is widely observed in the results. Compared to fair node classifiers, GCN shows promising results with improved fairness and accuracy. Most fairness issues with missing data are associated with sensitive class imbalance. Different imputation methods also affect fairness and accuracy. The exact effect on fairness and accuracy is subject to the used classifier, imputation technique, and the specific dataset. This research offers fertile observations for future work because there is very limited progress in this direction.


\bibliographystyle{splncs04}
\bibliography{references}

\begin{thebibliography}{10}
\providecommand{\url}[1]{\texttt{#1}}
\providecommand{\urlprefix}{URL }
\providecommand{\doi}[1]{https://doi.org/#1}

\bibitem{agarwal2021towards}
Agarwal, C., Lakkaraju, H., Zitnik, M.: Towards a unified framework for fair
  and stable graph representation learning. In: Uncertainty in Artificial
  Intelligence. pp. 2114--2124. PMLR (2021)

\bibitem{ali2021ssag}
Ali, S., Ahmad, M., Beg, M.A., Khan, I., Zaman, A., Khan, M.A.: Ssag:
  Summarization and sparsification of attributed graphs. arXiv preprint
  arXiv:2109.15111  (2022)

\bibitem{ali2021predicting}
Ali, S., Shakeel, M.H., Khan, I., Faizullah, S., Khan, M.A.: Predicting
  attributes of nodes using network structure. ACM Transactions on Intelligent
  Systems and Technology (TIST)  \textbf{12}(2),  1--23 (2021)

\bibitem{bakker2020fair}
Bakker, M., Vald{\'e}s, H.R., D~Tu, P., Gummadi, K.P., Varshney, K.R., Weller,
  A., Pentland, A.S.: Fair enough: Improving fairness in budget-constrained
  decision making using confidence thresholds  (2020)

\bibitem{beutel2017data}
Beutel, A., Chen, J., Zhao, Z., Chi, E.H.: Data decisions and theoretical
  implications when adversarially learning fair representations. arXiv preprint
  arXiv:1707.00075  (2017)

\bibitem{bruna2013spectral}
Bruna, J., Zaremba, W., Szlam, A., LeCun, Y.: Spectral networks and locally
  connected networks on graphs. arXiv preprint arXiv:1312.6203  (2013)

\bibitem{buyl2020debayes}
Buyl, M., De~Bie, T.: Debayes: a bayesian method for debiasing network
  embeddings. In: International Conference on Machine Learning. pp. 1220--1229.
  PMLR (2020)

\bibitem{chiang2019cluster}
Chiang, W.L., Liu, X., Si, S., Li, Y., Bengio, S., Hsieh, C.J.: Cluster-gcn: An
  efficient algorithm for training deep and large graph convolutional networks.
  In: Proceedings of the 25th ACM SIGKDD international conference on knowledge
  discovery \& data mining. pp. 257--266 (2019)

\bibitem{creager2019flexibly}
Creager, E., Madras, D., Jacobsen, J.H., Weis, M., Swersky, K., Pitassi, T.,
  Zemel, R.: Flexibly fair representation learning by disentanglement. In:
  International conference on machine learning. pp. 1436--1445. PMLR (2019)

\bibitem{dai2021say}
Dai, E., Wang, S.: Say no to the discrimination: Learning fair graph neural
  networks with limited sensitive attribute information. In: International
  Conference on Web Search and Data Mining. pp. 680--688 (2021)

\bibitem{defferrard2016convolutional}
Defferrard, M., Bresson, X., Vandergheynst, P.: Convolutional neural networks
  on graphs with fast localized spectral filtering. Advances in neural
  information processing systems  \textbf{29} (2016)

\bibitem{dong2022edits}
Dong, Y., Liu, N., Jalaian, B., Li, J.: Edits: Modeling and mitigating data
  bias for graph neural networks. In: Proceedings of the ACM Web Conference
  2022. pp. 1259--1269 (2022)

\bibitem{dwork2012fairness}
Dwork, C., Hardt, M., Pitassi, T., Reingold, O., Zemel, R.: Fairness through
  awareness. In: Proceedings of the 3rd innovations in theoretical computer
  science conference. pp. 214--226 (2012)

\bibitem{edwards2015censoring}
Edwards, H., Storkey, A.: Censoring representations with an adversary. arXiv
  preprint arXiv:1511.05897  (2015)

\bibitem{fan2021fair}
Fan, W., Liu, K., Xie, R., Liu, H., Xiong, H., Fu, Y.: Fair graph auto-encoder
  for unbiased graph representations with wasserstein distance. In: 2021 IEEE
  International Conference on Data Mining (ICDM). pp. 1054--1059. IEEE (2021)

\bibitem{garg2020fairness}
Garg, P., Villasenor, J., Foggo, V.: Fairness metrics: A comparative analysis.
  In: 2020 IEEE International Conference on Big Data (Big Data). pp.
  3662--3666. IEEE (2020)

\bibitem{giles1998citeseer}
Giles, C.L., Bollacker, K.D., Lawrence, S.: Citeseer: An automatic citation
  indexing system. In: Proceedings of the third ACM conference on Digital
  libraries. pp. 89--98 (1998)

\bibitem{grover2016node2vec}
Grover, A., Leskovec, J.: node2vec: Scalable feature learning for networks. In:
  Proceedings of the 22nd ACM SIGKDD international conference on Knowledge
  discovery and data mining. pp. 855--864 (2016)

\bibitem{hamaguchi2017knowledge}
Hamaguchi, T., Oiwa, H., Shimbo, M., Matsumoto, Y.: Knowledge transfer for
  out-of-knowledge-base entities: A graph neural network approach. arXiv
  preprint arXiv:1706.05674  (2017)

\bibitem{hamilton2017inductive}
Hamilton, W., Ying, Z., Leskovec, J.: Inductive representation learning on
  large graphs. Advances in neural information processing systems  \textbf{30}
  (2017)

\bibitem{hardt2016equality}
Hardt, M., Price, E., Srebro, N.: Equality of opportunity in supervised
  learning. Advances in neural information processing systems  \textbf{29}
  (2016)

\bibitem{henaff2015deep}
Henaff, M., Bruna, J., LeCun, Y.: Deep convolutional networks on
  graph-structured data. arXiv preprint arXiv:1506.05163  (2015)

\bibitem{khajehnejad2022crosswalk}
Khajehnejad, A., Khajehnejad, M., Babaei, M., Gummadi, K.P., Weller, A.,
  Mirzasoleiman, B.: Crosswalk: fairness-enhanced node representation learning.
  In: AAAI Conference on Artificial Intelligence. vol.~36, pp. 11963--11970
  (2022)

\bibitem{khosla2019comparative}
Khosla, M., Setty, V., Anand, A.: A comparative study for unsupervised network
  representation learning. IEEE Transactions on Knowledge and Data Engineering
  \textbf{33}(5),  1807--1818 (2019)

\bibitem{kipf2016semi}
Kipf, T.N., Welling, M.: Semi-supervised classification with graph
  convolutional networks. arXiv preprint arXiv:1609.02907  (2016)

\bibitem{li2019misgan}
Li, S.C.X., Jiang, B., Marlin, B.: Misgan: Learning from incomplete data with
  generative adversarial networks. arXiv preprint arXiv:1902.09599  (2019)

\bibitem{little2019statistical}
Little, R.J., Rubin, D.B.: Statistical analysis with missing data, vol.~793.
  John Wiley \& Sons (2019)

\bibitem{louizos2015variational}
Louizos, C., Swersky, K., Li, Y., Welling, M., Zemel, R.: The variational fair
  autoencoder. arXiv preprint arXiv:1511.00830  (2015)

\bibitem{lovedeep2018multiple}
Lovedeep, G., Mida, W.K.: Multiple imputation using denoising autoencoders. In:
  Pacific-Asia Conference on Knowledge Discovery and Data Mining. pp. 260--272
  (2018)

\bibitem{martinez2019fairness}
Mart{\'\i}nez-Plumed, F., Ferri, C., Nieves, D., Hern{\'a}ndez-Orallo, J.:
  Fairness and missing values. arXiv preprint arXiv:1905.12728  (2019)

\bibitem{perozzi2014deepwalk}
Perozzi, B., Al-Rfou, R., Skiena, S.: Deepwalk: Online learning of social
  representations. In: Proceedings of the 20th ACM SIGKDD international
  conference on Knowledge discovery and data mining. pp. 701--710 (2014)

\bibitem{Rahman2019FairwalkTF}
Rahman, T.A., Surma, B., Backes, M., Zhang, Y.: Fairwalk: Towards fair graph
  embedding. In: IJCAI (2019)

\bibitem{rajkomar2018ensuring}
Rajkomar, A., Hardt, M., Howell, M.D., Corrado, G., Chin, M.H.: Ensuring
  fairness in machine learning to advance health equity. Annals of internal
  medicine  \textbf{169}(12),  866--872 (2018)

\bibitem{spinelli2020missing}
Spinelli, I., Scardapane, S., Uncini, A.: Missing data imputation with
  adversarially-trained graph convolutional networks. Neural Networks
  \textbf{129},  249--260 (2020)

\bibitem{tang2020transferring}
Tang, X., Li, Y., Sun, Y., Yao, H., Mitra, P., Wang, S.: Transferring
  robustness for graph neural network against poisoning attacks. In:
  Proceedings of the 13th international conference on web search and data
  mining. pp. 600--608 (2020)

\bibitem{tang2020investigating}
Tang, X., Yao, H., Sun, Y., Wang, Y., Tang, J., Aggarwal, C., Mitra, P., Wang,
  S.: Investigating and mitigating degree-related biases in graph convoltuional
  networks. In: International Conference on Information \& Knowledge
  Management. pp. 1435--1444 (2020)

\bibitem{troyanskaya2001missing}
Troyanskaya, O., Cantor, M., Sherlock, G., Brown, P., Hastie, T., Tibshirani,
  R., Botstein, D., Altman, R.B.: Missing value estimation methods for dna
  microarrays. Bioinformatics  \textbf{17}(6),  520--525 (2001)

\bibitem{velickovic2017graph}
Velickovic, P., Cucurull, G., Casanova, A., Romero, A., Lio, P., Bengio, Y.:
  Graph attention networks. stat  \textbf{1050}, ~20 (2017)

\bibitem{yang2019categorical}
Yang, Y., Wu, Z., Tresp, V., Fasching, P.A.: Categorical ehr imputation with
  generative adversarial nets. In: 2019 IEEE International Conference on
  Healthcare Informatics (ICHI). pp. 1--10. IEEE (2019)

\bibitem{yao2019graph}
Yao, L., Mao, C., Luo, Y.: Graph convolutional networks for text
  classification. In: Proceedings of the AAAI conference on artificial
  intelligence. vol.~33, pp. 7370--7377 (2019)

\bibitem{ying2018graph}
Ying, R., He, R., Chen, K., Eksombatchai, P., Hamilton, W.L., Leskovec, J.:
  Graph convolutional neural networks for web-scale recommender systems. In:
  Proceedings of the 24th ACM SIGKDD international conference on knowledge
  discovery \& data mining. pp. 974--983 (2018)

\bibitem{you2020handling}
You, J., Ma, X., Ding, Y., Kochenderfer, M.J., Leskovec, J.: Handling missing
  data with graph representation learning. Advances in Neural Information
  Processing Systems  \textbf{33},  19075--19087 (2020)

\bibitem{zhang2021multi}
Zhang, X., Zhang, L., Jin, B., Lu, X.: A multi-view confidence-calibrated
  framework for fair and stable graph representation learning. In: 2021 IEEE
  International Conference on Data Mining (ICDM). pp. 1493--1498. IEEE (2021)

\bibitem{zhang2021assessing}
Zhang, Y., Long, Q.: Assessing fairness in the presence of missing data.
  Advances in neural information processing systems  \textbf{34},  16007--16019
  (2021)

\bibitem{Fairness_in_missing_data}
Zhang, Y., Long, Q.: Fairness in missing data imputation. arXiv preprint
  arXiv:2110.12002  (2021)

\bibitem{zhao2020semi}
Zhao, T., Tang, X., Zhang, X., Wang, S.: Semi-supervised graph-to-graph
  translation. In: International Conference on Information \& Knowledge
  Management. pp. 1863--1872 (2020)

\end{thebibliography}
\end{document}